\documentclass[11pt,a4paper]{article}

\usepackage[T1]{fontenc}
\usepackage[utf8]{inputenc}

\usepackage{mathptmx}
\usepackage[scaled=0.92]{helvet}

\usepackage{amsmath}
\usepackage{amssymb}

\usepackage{xpeng-brand}

\usepackage{url}
\usepackage{graphicx}
\usepackage{booktabs}
\usepackage{array}
\usepackage{tabularx}
\usepackage{multirow}
\usepackage{nicefrac}
\usepackage{pifont}
\usepackage{wrapfig}
\usepackage[numbers,sort&compress]{natbib}
\usepackage{hyperref}
\usepackage{microtype}
\usepackage{xspace}

\definecolor{EdgeBg}{HTML}{EDF7ED}
\definecolor{MidBg}{HTML}{E8F0FE}
\definecolor{MassiveBg}{HTML}{FFF3E0}
\definecolor{ClosedBg}{HTML}{F3E8FD}
\definecolor{FRtint}{HTML}{FDE8E8}
\definecolor{NRtint}{HTML}{E6F4F1}

\tcbset{
  promptbox/.style={
    colback=black!3,
    colframe=black!18,
    boxrule=0.35pt,
    arc=2pt,
    fonttitle=\rmfamily\bfseries\small,
    fontupper=\ttfamily\footnotesize,
    left=8pt, right=8pt, top=5pt, bottom=5pt,
    before skip=8pt, after skip=8pt,
  },
}

\newcommand{\nickname}{\textsc{Vigil}\xspace}
\newcommand{\cmark}{\ding{51}}
\newcommand{\xmark}{\ding{55}}

\xpengcorrespondence{Jie Chen}{chenj81@xiaopeng.com}

\title{Done, But Not Sure: Disentangling World Completion \\
  from Self-Termination in Embodied Agents}

\author{%
  Ying Chen, Lihuang Fang, Rui Jiang, Mingxu Wang, Zhifeng Gu, Lei Yi, Jie Chen$^{\dagger}$%
}
\date{}

\begin{document}

\begin{xpenghero}
\begin{abstract}
Standard embodied evaluations do not independently score whether an agent correctly commits to task completion at episode closure, a capacity we call \emph{terminal commitment}.
Behaviorally distinct failures---never completing the task, completing it but failing to stop, and reporting success without sufficient evidence---collapse into the same benchmark failure.
We introduce \textbf{\nickname}, an evaluation framework that makes terminal commitment independently measurable.
Under \nickname's default protocol, agents observe only egocentric RGB, receive no action-success signals, and must end each episode with a semantic report checked deterministically against hidden world state.
This yields two separate scores: world-state completion~(\textbf{W}) and benchmark success~(\textbf{B}), where \textbf{B} additionally requires a correct terminal report.
This decoupling makes four outcome categories distinguishable: missed execution, post-attainment drift, unsupported commitment, and verified success.
Across 20 models on 1{,}000 frozen episodes, systems with comparable \textbf{W} differ by up to 19.7\,pp in \textbf{B}: one model converts achieved states into correct reports, while another with near-identical execution drifts past the goal without closing.
An action-feedback intervention further tests the separation: execution-oriented signals improve \textbf{W} broadly, yet commitment failures persist in models that do not already ground terminal reports in the achieved state.
\nickname provides a protocol that makes terminal commitment independently visible and scorable.
\end{abstract}
\end{xpenghero}

\section{Introduction}
\label{sec:intro}


Embodied agents must not only take actions to complete a task, but also determine when the task has been completed and commit to that judgment.
This is non-trivial when agents operate under partial observability and must infer task progress from limited egocentric observations over time \citep{zhang2026theory, gao2025cubebench}, without action feedback or success signals \citep{brohan2023rt2, wang2025embodiedbench}.
For example, in a task that asks the agent to turn on a lamp, the agent may successfully turn it on yet continue navigating because it fails to recognize that the task is already complete.
Under current embodied evaluation, this outcome is often indistinguishable from never having completed the task.

\begin{figure}[t]
    \centering
    \includegraphics[width=1.0\columnwidth]{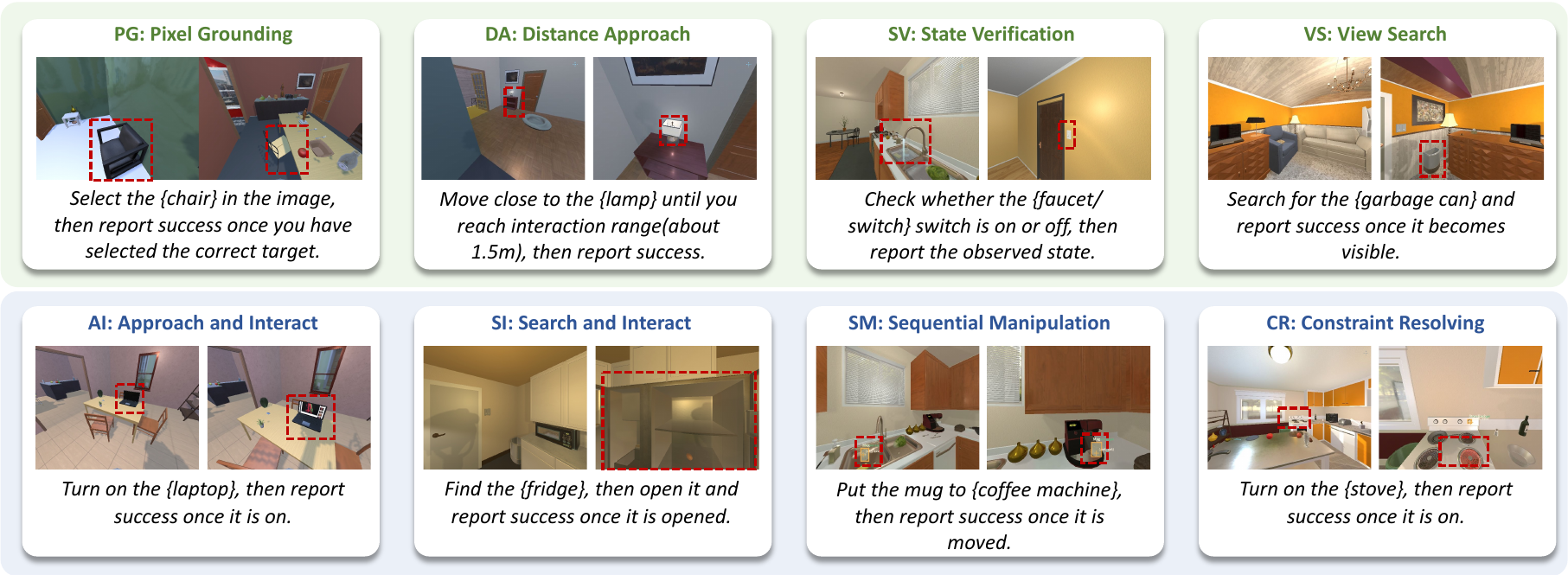}
    \caption{\textbf{Controlled evaluation protocol.} \nickname contains eight task families: a diagnostic tier (PG, DA, SV, VS) that isolates a single bottleneck, and a compositional tier (AI, SI, SM, CR) that combines them in multi-step interaction. All episodes use strict first-person observation and mandatory \texttt{report} termination.}
    \label{fig:benchmark}
\end{figure}


To our knowledge, no existing embodied benchmark cleanly separates task-closure failures from execution failures \citep{shridhar2020alfred, choi2024lota, khanna2024goatbench}. 
Recent work has improved per-skill and capability-level diagnostics \citep{li2024embodied, cheng2025embodiedeval, wang2025embodiedbench, peng2026nativeembodied}, yet overall model comparison is still summarized by aggregated task- or episode-level success metrics.
These metrics collapse behaviorally distinct cases into the same final outcome: the agent may fail to complete the task, complete it but fail to stop, or declare completion without sufficient evidence---failures with different causes and remedies that current metrics conflate.

We introduce \textbf{\nickname}, an evaluation framework that makes \textit{terminal commitment} independently measurable.
The design has three key elements.
First, agents observe only egocentric RGB, without privileged state, oracle progress signals, or action-success confirmation.
Second, each episode ends with a mandatory semantic \texttt{report} whose content is checked deterministically against the hidden world state.
Third, this yields two separate scores: world-state completion~(\textbf{W}) and benchmark success~(\textbf{B}), where \textbf{B} additionally requires a correct terminal commitment.
The no-feedback contract is essential: if agents received action-success signals, correct reports could be produced by echoing environment confirmation rather than by maintaining task-state judgment from observation.

This target differs from prior work on self-assessment and uncertainty in language agents \citep{kadavath2022language, ren2023knowno, huang2022innermonologue}.
Rather than inferring latent confidence or internal belief \cite{kadavath2022language}, we evaluate whether the agent expresses the correct task-state judgment at episode closure through a terminal report that can be checked against the hidden world state.
For state-verification tasks, where the agent must report \texttt{open}/\texttt{closed} or \texttt{on}/\texttt{off}, the correct terminal act is a categorical state judgment, not a binary decision to halt.
Consequently, a stop-only protocol cannot represent this class of failure, because the error lies in report content rather than in termination timing.

\nickname evaluates this dimension over eight task families that probe the conditions under which terminal judgment becomes difficult (Figure~\ref{fig:benchmark}), spanning target visibility, distance, state uncertainty, temporal dependency, and physical constraint.
A diagnostic tier (short budgets, single bottlenecks) isolates closure failures when execution is tractable, while a compositional tier (longer budgets, chained prerequisites) reveals when execution floors mask closure errors.
We additionally use an action-feedback intervention, modeled on proprioceptive signals available to physical robots, to test whether closure failures can be explained by upstream execution traps alone.

\paragraph{Findings.}
Across 20 multimodal systems spanning open-weight and closed-source frontier models (10 anchor models in the main text; full panel in Appendix~\ref{app:full_panel}), we find that execution and terminal commitment are empirically separable:
\begin{itemize}[leftmargin=*,nosep]
    \item \textbf{Structured closure-failure profiles.}
    Models exhibit distinct terminal behaviors, including premature false commitments (\textbf{FR}), chronic no-report exhaustion (\textbf{NR}), and selective reporting. These profiles are invisible under aggregate success and stable across prompt variants.
    \item \textbf{Execution floors mask closure failures.}
    On longer compositional tasks, execution often fails before terminal judgment can be meaningfully evaluated, compressing the observable gap and masking closure errors that the diagnostic tier isolates directly.
    \item \textbf{Execution feedback is not a universal fix.}
    A proprioceptive action-feedback intervention reduces execution traps broadly, but improves terminal reporting only for models whose terminal reports are already coupled to achieved task state.
\end{itemize}
Together, these results identify terminal commitment as a distinct failure dimension: it is empirically separable from execution, produces consistent step-level patterns, and is not uniformly repaired by improving execution alone.
To the best of our knowledge, \nickname provides the first evaluation protocol that makes this dimension independently measurable, enabling targeted diagnosis and comparison of terminal judgment across embodied systems.

\section{Related Work}
\label{sec:related_work}

\nickname intersects three lines of work: embodied evaluation, self-assessment and confidence-related control, and belief-oriented embodied reasoning.
Its scientific target is distinct: whether the agent can correctly judge and report its achieved task state at episode closure, scored independently of execution success.
Table~\ref{tab:benchmark_comparison} summarizes how existing settings compare along this axis.

\begin{table}[t]
\centering
\caption{Representative embodied evaluation settings and whether they make terminal task-state judgment externally scorable. \nickname makes agent-side terminal commitment independently scorable under a no-feedback native-control contract.}
\resizebox{\textwidth}{!}{
\begin{tabular}{lcccccccc}
\toprule
\textbf{Benchmark} & \textbf{Size} & \textbf{Task Level} & \textbf{Multimodal} & \textbf{Native} & \textbf{Fine-Grained} & \textbf{Decoupled} & \textbf{Active Termination} & \textbf{No Feedback} \\
\midrule
ALFRED \citep{shridhar2020alfred} & 3,062 & High & \cmark & \xmark & \xmark & \xmark & Stop & \xmark \\
ALFWorld \citep{shridhar2021alfworld} & 274 & High & \xmark & \xmark & \xmark & \xmark & \xmark & \xmark \\
VLMbench \citep{zheng2022vlmbench} & 4,760 & Low & \cmark & \cmark & \xmark & \xmark & \xmark & \xmark \\
BEHAVIOR-1K \citep{li2023behavior1k} & 1{,}000 & High & \cmark & \xmark & \xmark & \xmark & \xmark & \xmark \\
LoTa-Bench \citep{choi2024lota} & 308 & High & \xmark & \cmark & \xmark & \xmark & Stop & \xmark \\
GOAT-Bench \citep{khanna2024goatbench} & 3,919 & Low & \cmark & \cmark & \xmark & \xmark & Stop & \xmark \\
Embodied Agent Interface \citep{li2024embodied} & 438 & High & \xmark & \xmark & \cmark & \xmark & \xmark & \xmark \\
EmbodiedBench \citep{wang2025embodiedbench} & 1,128 & High\&Low & \cmark & \xmark & \cmark & \xmark & Stop & \xmark \\
EmbodiedEval \citep{cheng2025embodiedeval} & 328 & High & \cmark & \xmark & \xmark & \xmark & \xmark & \xmark \\
NativeEmbodied \citep{peng2026nativeembodied} & 1,085 & High\&Low & \cmark & \cmark & \cmark & Task & \xmark & \xmark \\
\midrule
\textbf{\nickname (Ours)} & 1{,}000 & High\&Low & \cmark & \cmark & \cmark & Task+Report & Report & \cmark \\
\bottomrule
\end{tabular}
}
\label{tab:benchmark_comparison}
\end{table}

\noindent
Column definitions: \emph{Native}---agent acts through natural skill calls rather than privileged API commands.
\emph{Fine-Grained}---per-skill diagnostics beyond aggregate success.
\emph{Active Termination}---``Stop'': bare stop action absorbed into task success; ``Report'': semantic terminal judgment independently scored; \xmark: evaluator-side termination only.
\emph{Decoupled}---``Task'': per-skill decomposition without independent report scoring; ``Task+Report'': adds deterministically scored terminal commitment.
\emph{No Feedback}---no external confirmation of task progress, goal completion, or action success.

\paragraph{Embodied Evaluation Protocols.}
Household instruction-following benchmarks define long-horizon tasks evaluated by terminal success \citep{shridhar2020alfred, shridhar2021alfworld, padmakumar2022teach, kim2024realfred}.
Subsequent work extends the setting across simulation fidelity \citep{li2023behavior1k}, LLM-based planning \citep{choi2024lota}, navigation \citep{khanna2024goatbench}, and compositional manipulation \citep{zheng2022vlmbench}.
Recent suites add per-skill diagnostics \citep{li2024embodied, wang2025embodiedbench, cheng2025embodiedeval, peng2026nativeembodied}, improving attribution across perception, navigation, and manipulation.
The key gap remains: terminal success is an evaluator-side predicate over world state, not an agent-side judgment that is independently scored.
Execution failures and closure failures are therefore behaviorally conflated.

\paragraph{Self-Assessment, Confidence, and Termination Decisions.}
Several benchmarks include a \texttt{stop} action \citep{shridhar2020alfred, choi2024lota, khanna2024goatbench}, but stopping is a control primitive absorbed into task success without independent evaluation.
A parallel literature asks whether language models possess calibrated self-knowledge \citep{kadavath2022language} and whether confidence signals can trigger help-seeking or replanning \citep{ren2023knowno, huang2022innermonologue}.
\nickname targets a different observable: not latent confidence, but whether the agent produces a semantically correct terminal report that can be verified against hidden world state.

\paragraph{Belief Under Embodied Interaction.}
Spatial intelligence in vision-language models is increasingly studied beyond static images, spanning metric reasoning \citep{chen2024spatialvlm, cheng2024spatialrgpt}, perspective-taking \citep{omnispatial25, yang2025mmsi, wang2025spatial457}, and robotics-oriented grounding \citep{yuan2024robopoint, song2025robospatial, yang2025cambrians, zhou2025vlm4d}.
These settings show that embodied interaction requires building and updating representations from partial observations---a capacity that remains difficult for current models \citep{yang2024think, yin2025mental}.
Recent work makes this explicit: Theory of Space \citep{zhang2026theory} asks whether foundation models construct spatial beliefs through active exploration; CubeBench \citep{gao2025cubebench} diagnoses interactive spatial reasoning under partial observation.
\nickname shares this belief-under-interaction perspective but targets a different output: task-state judgment at closure, not spatial belief during exploration.

\section{Benchmark Design}
\label{sec:method}

\nickname separates two outcomes that standard embodied evaluation conflates: whether the agent changed the world correctly, and whether it issued a correct terminal report about that change.
This requires controlled task families with hidden goal predicates, a no-feedback interaction contract, and a deterministic scoring rule that evaluates world state and report content independently.

\subsection{Task Families}
\label{sec:taxonomy}

Each frozen episode specifies a task instruction, a family label, step and invalid-action budgets, and a hidden success condition $\mathcal{G}$ checked against the simulator state.
The benchmark contains 1{,}000 episodes across eight balanced families (125 each), organized as controlled probes over the factors that shape task-state judgment under interaction (Figure~\ref{fig:benchmark}).

Rather than decomposing by domain-specific subtasks---as in recent skill-diagnostic benchmarks \citep{wang2025embodiedbench, peng2026nativeembodied}---we decompose along atomic perceptual-motor capacities and progressively compose them.
This makes attribution precise: when a model fails on a composed task but succeeds on its constituents, the bottleneck lies in composition or terminal judgment, not in the constituent skills.

The \emph{diagnostic tier} isolates single bottlenecks with short budgets (5--20 steps):
\textbf{PG} (pixel grounding)---click the correct visible object;
\textbf{DA} (distance approach)---navigate to a visible target;
\textbf{VS} (view search)---find a non-visible target through active exploration;
\textbf{SV} (state verification)---report the categorical state of a visible object without physical interaction.
SV provides the purest test of terminal commitment: world-state completion exceeds 80\% for most models, so failures in \textbf{B} directly expose judgment errors.

The \emph{compositional tier} chains these capacities under longer budgets (25--40 steps):
\textbf{AI} (approach-and-interact\,=\,DA\,+\,interaction);
\textbf{SI} (search-and-interact\,=\,VS\,+\,DA\,+\,interaction);
\textbf{SM} (sequential manipulation---multi-step pick-and-place);
\textbf{CR} (constraint resolving---obstacle removal before interaction).
Full family specifications are in Appendix~\ref{app:task_families}.

\subsection{Native-Control Contract}
\label{sec:environment}

Episodes run in AI2-THOR~\citep{kolve2017ai2thor} with ProcTHOR~\citep{deitke2022procthor} houses.
The agent receives a single egocentric RGB frame per step and acts through four skills: locomotion (\texttt{navigate}, \texttt{look}), pixel-grounded object interaction (\texttt{interact\_pixel}), and terminal reporting (\texttt{report}).
No privileged state or action-success feedback is provided: the agent receives no maps, semantic masks, absolute pose, or oracle progress signals.
State changes are visible only through subsequent RGB observations.

Following the dialogue-based native-control paradigm of recent embodied evaluation \citep{wang2025embodiedbench, peng2026nativeembodied}, the agent maintains a bounded dialogue history of up to 20 turns (one turn per action step)---including any agent-generated \texttt{thought} or \texttt{cognitive\_state} fields---as the only cross-step memory.
The 20-turn bound is chosen to match the step budgets of our longest compositional tasks (25--40 steps) while remaining within the reliable context-utilization range of current multimodal models.

This contract is intentionally sparse: terminal judgments must be formed from the agent's own partial-observation interaction history rather than from evaluator-side success signals.

\begin{figure}[t]
    \centering
    \includegraphics[width=1.0\columnwidth]{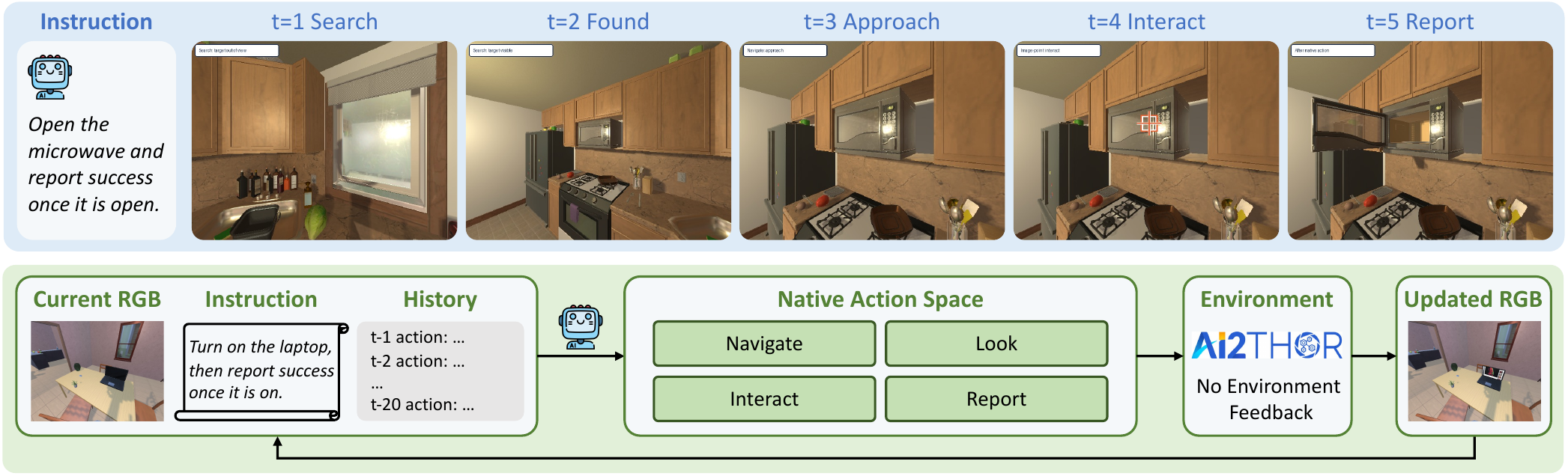}
    \caption{\textbf{Evaluation pipeline.} 
    The top row illustrates an example trajectory.
    The bottom row summarizes the per-step interface: the agent acts from only the current egocentric RGB frame, the task instruction, and bounded dialogue history, without action-success or goal-completion feedback.
    A terminal \texttt{report} is evaluated against the hidden world-state condition by deterministic rules.}
    \label{fig:pipeline}
\end{figure}

The key mechanism is the terminal \texttt{report}: it carries a categorical \texttt{status} (e.g., \texttt{success}, \texttt{fail}, \texttt{on}/\texttt{off}, \texttt{open}/\texttt{closed}) and a short summary.
Unlike a bare \texttt{stop}, \texttt{report} records \emph{what} the agent judges the task state to be at termination.
This makes the terminal act externally scorable: the evaluator can check whether the reported judgment matches the hidden world state, rather than merely whether termination coincided with goal satisfaction.
This matters most in state-verification tasks, where the correct terminal act is a categorical state judgment rather than a binary decision to end.

\subsection{Disaggregated Outcomes}
\label{sec:termination}

We separate two outcomes at episode closure: whether the task condition is satisfied in the world, and whether the agent's terminal commitment correctly reflects that state.

For trajectory $\tau$, let $s_T$ be the hidden terminal state, $\mathcal{G}_{\mathrm{sem}}$ the primary semantic task-goal predicate, $a_T$ the terminal action, and $\texttt{report}_T$ the report content.
The report status is normalized to one of eight categorical values (\texttt{success}, \texttt{fail}, \texttt{on}, \texttt{off}, \texttt{open}, \texttt{closed}, \texttt{unsafe}, \texttt{invalid}) and matched against the hidden state by a deterministic rule.
For goal-completion tasks, \texttt{success} must coincide with $W\!=\!1$; for state-verification tasks, where $\mathcal{G}_{\mathrm{sem}}$ denotes the queried state predicate rather than a manipulation goal, the status must exactly match the hidden object state (e.g., \texttt{on} iff the target is toggled on).
We define:
\[
\begin{aligned}
    W(\tau) &= \mathbb{I}[s_T \models \mathcal{G}_{\mathrm{sem}}], \\
B(\tau) &= \mathbb{I}\!\left[
    s_T \models \mathcal{G}_{\mathrm{sem}}
    \;\land\;
    a_T = \texttt{report}
    \;\land\;
    \texttt{match}(\texttt{report}_T, s_T)
\right].
\end{aligned}
\]
$W$ captures world-state completion at termination under the primary semantic task predicate.
$B$ captures benchmark success, requiring not only the correct semantic world state but also a matching terminal commitment.
If an episode ends by budget exhaustion or invalid-action limit before \texttt{report}, then $B=0$ regardless of the world state.
The gap $\Delta = W - B$ isolates achieved task states not converted into benchmark success through correct terminal commitment.

Importantly, $\Delta$ is a decomposition handle rather than a standalone reliability measure: a small gap may reflect either reliable state-coupled commitment or aggressive reporting that coincidentally aligns with low $W$.
We therefore analyze $\Delta$ jointly with deterministic closure-failure labels:
\textbf{False reports} (\textbf{FR}) occur when the report claims a status unsupported by the hidden state.
\textbf{No-report exhaustion} (\textbf{NR}) occurs when the step budget expires without a \texttt{report}.
\textbf{Invalid-action-limit} terminations (\textbf{IL}) occur when cumulative malformed actions exceed the family budget.
Appendix~\ref{app:failure_taxonomy} summarizes the reported labels used in the figures and tables.

\paragraph{Interpretation Note.}
VIGIL does not claim that every failure labeled by $B=0$ uniquely identifies an internal cognitive cause.
Rather, it provides an externally testable decomposition of closure behavior: whether the agent achieved the relevant task state, whether it issued a terminal judgment, and whether that judgment matched the achieved world state.
This decomposition allows closure failures to be compared systematically across models and interventions.

\section{Experiments}
\label{sec:experiments}

All experiments use the native-control contract from \S\ref{sec:method}: egocentric RGB input, bounded dialogue history, no environment feedback, and mandatory \texttt{report} termination.
Main-text world-state results (\textbf{W}) use the primary semantic goal predicate; dual-metric scoring details are in Appendix~\ref{app:scoring_details}.
We evaluate a frozen 1{,}000-episode set (eight families, 125 each) on 10 anchor models spanning closed-source frontier (Gemini-3.1-Pro \citep{deepmind2026gemini31pro}, Doubao-Seed-1.8 \citep{guo2025seed15vl}, GPT-5.4 \citep{openai2026gpt54}, Claude-Sonnet-4 \citep{anthropic2025claude4}), open general VLMs (Qwen3.6-27B \citep{qwenteam2026qwen3627b}, Qwen3.5-27B \citep{qwenteam2026qwen35}---both served with thinking enabled---Qwen3-VL-32B \citep{bai2025qwen3} in Instruct and Thinking variants, and InternVL3.5-38B \citep{wang2025internvl35}), and embodied-tuned systems (MiMo-Embodied-7B \citep{xiaomi2025mimo}).
This spread tests whether terminal commitment failures are specific to a model class or appear broadly across architectures, scales, and training objectives; the primary 20-model comparison panel is in Appendix~\ref{app:full_panel}.

The empirical goal is not only to compare benchmark success, but to characterize how embodied models fail at episode closure.
We ask three questions: do execution and terminal commitment separate empirically (\S\ref{sec:cross_section})? Do models exhibit distinct closure-failure profiles with interpretable step-level signatures (\S\ref{sec:terminal_commitment})? Are these closure failures explained entirely by upstream execution traps (\S\ref{sec:feedback})?

\subsection{Cross-Model Outcomes: Execution and Terminal Commitment Separate}
\label{sec:cross_section}

\begin{table*}[t]
\centering
\caption{%
\textbf{Per-family outcomes for 10 anchor models under native control.}
Each cell reports \textbf{W}/\textbf{B} (\%), where \textbf{W}\,=\,world-state completion and \textbf{B}\,=\,benchmark success.
$\boldsymbol{\Delta}$\,=\,$W\!-\!B$ gap (pp).
Diagnostic: PG\,=\,pixel grounding, DA\,=\,distance approach, VS\,=\,view search, SV\,=\,state verification.
Compositional: AI\,=\,approach-and-interact, SI\,=\,search-and-interact, SM\,=\,sequential manipulation, CR\,=\,constraint resolving.
Anchor set shared with all subsequent analyses; primary 20-model comparison panel in Appendix~\ref{app:full_panel}. \texttt{(T)} marks thinking serving mode; Qwen3.6-27B and Qwen3.5-27B are served with thinking enabled and not additionally marked.%
}
\label{tab:global_cross_section}
\scriptsize
\setlength{\tabcolsep}{3.0pt}
\renewcommand{\arraystretch}{1.18}
\begin{tabularx}{\textwidth}{@{}>{\raggedright\arraybackslash}p{2.75cm}
  *{2}{>{\centering\arraybackslash}X}
  *{8}{>{\centering\arraybackslash}X}@{}}
\toprule
\multirow{2}{*}[-0.3ex]{\textbf{Model}}
  & \textbf{All}
  & \multirow{2}{*}[-0.3ex]{$\boldsymbol{\Delta}$}
  & \multicolumn{4}{c}{\textbf{Diagnostic Probes}}
  & \multicolumn{4}{c@{}}{\textbf{Compositional Tasks}} \\
\cmidrule(lr){2-2}\cmidrule(lr){4-7}\cmidrule(l){8-11}
  & \textbf{W/B} & & \textbf{PG} & \textbf{DA} & \textbf{VS} & \textbf{SV}
  & \textbf{AI} & \textbf{SI} & \textbf{SM} & \textbf{CR} \\
\midrule
Gemini-3.1-Pro\rlap{$^\dagger$} & \textbf{57.7}/\textbf{56.4} & \textbf{1.3} & \textbf{92.0}/\textbf{92.0} & 52.8/52.0 & \textbf{71.2}/\textbf{71.2} & \textbf{95.2}/\textbf{90.4} & \textbf{62.4}/\textbf{60.8} & \textbf{26.4}/\textbf{25.6} & \textbf{43.2}/\textbf{42.4} & \textbf{18.4}/\textbf{16.8} \\
Qwen3.6-27B       & 46.7/37.8 & 8.9 & 87.2/76.8 & \textbf{56.8}/\textbf{55.2} & 47.2/44.0 & 93.6/80.8 & 54.4/28.8 & 12.8/7.2 & 14.4/6.4 & 7.2/3.2 \\
Doubao-Seed-1.8\rlap{$^\dagger$} & 40.4/37.1 & 3.3 & 76.8/75.2 & 44.8/41.6 & 29.6/29.6 & 94.4/80.0 & 42.4/37.6 & 5.6/4.8 & 19.2/17.6 & 10.4/10.4 \\
Qwen3.5-27B       & 42.0/36.2 & 5.8 & 84.0/80.8 & 41.6/39.2 & 48.0/44.8 & 93.6/85.6 & 36.8/19.2 & 15.2/8.0 & 10.4/9.6 & 6.4/2.4 \\
Qwen3-VL-32B \texttt{(T)} & 30.7/26.4 & 4.3 & 70.4/68.0 & 10.4/10.4 & 43.2/41.6 & 93.6/81.6 & 15.2/4.8 & 2.4/0.8 & 5.6/3.2 & 4.8/0.8 \\
GPT-5.4\rlap{$^\dagger$} & 29.2/24.7 & 4.5 & 30.4/30.4 & 24.8/24.8 & 64.8/62.4 & 92.8/72.0 & 10.4/0.8 & 4.0/1.6 & 4.8/4.0 & 1.6/1.6 \\
MiMo-Embodied-7B  & 29.5/21.3 & 8.2 & 52.0/34.4 & 37.6/24.0 & 31.2/26.4 & \textbf{95.2}/74.4 & 14.4/7.2 & 0.8/0.8 & 3.2/1.6 & 1.6/1.6 \\
Claude-Sonnet-4\rlap{$^\dagger$} & 25.2/21.1 & 4.1 & 38.4/35.2 & 17.6/16.8 & 65.6/64.8 & 73.6/45.6 & 4.8/4.8 & 0.0/0.0 & 1.6/1.6 & 0.0/0.0 \\
Qwen3-VL-32B      & 39.2/18.9 & 20.3 & 68.8/20.0 & 44.0/14.4 & 65.6/48.8 & 82.4/51.2 & 34.4/8.0 & 5.6/0.8 & 7.2/5.6 & 5.6/2.4 \\
InternVL3.5-38B   & 39.9/17.4 & 22.5 & 79.2/16.0 & 57.6/18.4 & 50.4/36.8 & 86.4/55.2 & 31.2/7.2 & 7.2/0.0 & 5.6/4.8 & 1.6/0.8 \\
\bottomrule
\end{tabularx}
\renewcommand{\arraystretch}{1.0}
\vskip 0.05in
{\scriptsize $^\dagger$Closed-source API models. Identifiers: \texttt{gemini-3.1-pro-preview}, \texttt{doubao-seed-1.8}, \texttt{gpt-5.4}, \texttt{claude-sonnet-4-20250514}.}
\vskip -0.10in
\end{table*}

Table~\ref{tab:global_cross_section} summarizes per-family outcomes for the 10 anchor models used throughout all analyses.

\paragraph{Execution and terminal commitment separate empirically.}
Comparable levels of world-state completion can produce sharply different benchmark success.
Doubao-Seed-1.8 and InternVL3.5-38B reach nearly identical \textbf{W} (40.4\% vs.\ 39.9\%), yet differ by 19.7\,pp in \textbf{B} (37.1\% vs.\ 17.4\%); their sharply different closure gaps ($\Delta\!=\!3.3$ vs.\ 22.5\,pp) explain this benchmark-success divergence.
A related decoupling appears between Qwen3.6-27B (46.7\%/37.8\%) and InternVL3.5-38B (39.9\%/17.4\%).
The reverse also holds: Qwen3-VL-32B \texttt{(T)} has lower \textbf{W} than its instruct counterpart (30.7\% vs.\ 39.2\%) but higher \textbf{B} (26.4\% vs.\ 18.9\%), showing that higher execution does not guarantee higher benchmark success.

The factorial task structure enables further attribution: because diagnostic families isolate single capacities, we can localize where closure breaks down in composed tasks.
Qwen3.6-27B performs well on PG (87.2\%) and reasonably on DA (56.8\%) individually, but on AI attains 54.4\% \textbf{W} and only 28.8\% \textbf{B}, suggesting that the composed-task bottleneck is not merely constituent navigation or grounding but also conversion of successful interaction into terminal commitment.
For open-weight configurations with paired robustness runs, prompt-sensitivity and repeated-run checks show that the main closure profiles are stable across prompt variants and reruns (Appendix~\ref{app:robustness_checks}).

\paragraph{Compositional tasks reveal where execution floors mask closure failures.}
Large $W$--$B$ separations concentrate in diagnostic probes and the simplest compositional task (AI), where execution remains high enough for closure behavior to be observable.
On deeper compositional tasks (SI, SM, CR), the gap shrinks because successful world-state attainment itself becomes rare---since $\Delta$ is bounded by \textbf{W}, low execution compresses the observable signature of closure failure.
The compositional tier therefore shows when execution becomes the dominant bottleneck and masks the closure failures that shorter episodes isolate directly.

\subsection{Terminal Commitment: Structured Failures and Step-Level Evidence}
\label{sec:terminal_commitment}

We now decompose terminal failure modes and examine their step-level signatures to determine whether closure failures have interpretable behavioral structure or are merely stochastic noise.

\begin{figure}[t]
    \centering
    \includegraphics[width=0.9\columnwidth]{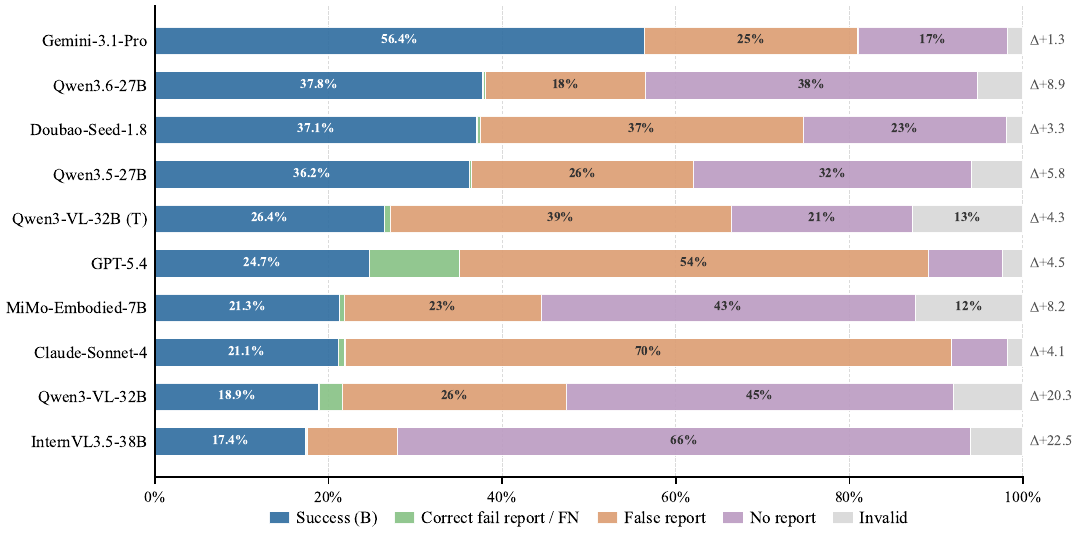}
    \caption{\textbf{Episode outcome partition for 10 anchor models} (sorted by \textbf{B}).
    FR-heavy profiles (Claude-Sonnet-4, GPT-5.4) and NR-heavy profiles (InternVL3.5-38B, Qwen3-VL-32B) are immediately distinguishable despite comparable aggregate success.
    Label definitions are in Appendix~\ref{app:failure_taxonomy}; counterfactual report-policy analysis is in Appendix~\ref{app:report_policy_baselines}.}
    \label{fig:failure_distribution}
\end{figure}

\paragraph{False commitments and missed commitments define distinct closure regimes.}
Figure~\ref{fig:failure_distribution} shows that terminal failure takes two dominant forms: unsupported commitment (FR---the agent issues a terminal report whose status is contradicted by the hidden world state) and missed terminal commitment (NR---the agent exhausts the budget without issuing a report).
Within $W\!=\!1$ episodes, NR corresponds to post-attainment drift.
FR-heavy models (GPT-5.4 at 54.0\%, Claude-Sonnet-4 at 69.9\%) frequently issue terminal judgments unsupported by the achieved world state.
NR-heavy models (InternVL3.5-38B at 66.1\%, Qwen3-VL-32B at 44.6\%, MiMo-Embodied-7B at 43.1\%) often exhaust the budget without a terminal report; among episodes where the world goal is reached, this appears as failure to convert task-state attainment into terminal commitment.
Qwen3-VL-32B combines both modes---substantial FR and NR---producing a $\Delta$ of 20.3\,pp despite nontrivial \textbf{W}, demonstrating that the two failure regimes are not mutually exclusive.

\paragraph{State verification makes semantic terminal judgment indispensable.}
SV episodes provide the clearest test of report content because the object state is already set at episode start and no physical interaction is required beyond observation: \textbf{W} exceeds 80\% for 9 of 10 anchors (Table~\ref{tab:global_cross_section}).
Yet \textbf{B} drops by 5--31\,pp, with the largest gaps in Claude-Sonnet-4 (73.6$\to$45.6), GPT-5.4 (92.8$\to$72.0), and InternVL3.5-38B (86.4$\to$55.2).
Because the correct terminal output is categorical (\texttt{on}/\texttt{off}, \texttt{open}/\texttt{closed}), a stop-only protocol can observe whether the agent terminates but not whether its terminal state judgment is semantically correct---semantic terminal evaluation is required to capture this class of embodied judgment error.

\begin{figure*}[t]
    \centering
    \includegraphics[width=\textwidth]{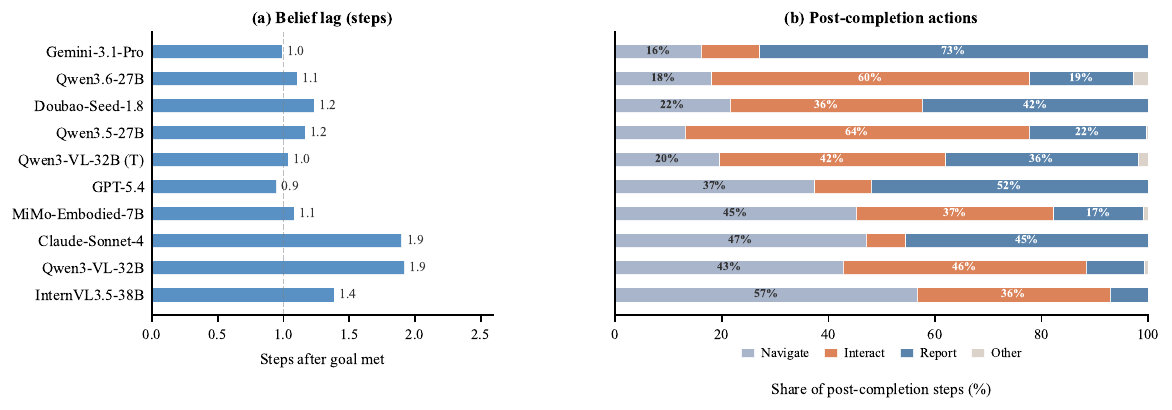}
    \caption{\textbf{Terminal-commitment profiles} (sorted by \textbf{B} descending).
    (a)~Mean belief lag: steps between first world-goal satisfaction and the correct terminal report.
    When agents do report correctly, they do so within 0.9--1.9\,steps of that event (panel~(a) rounds each model to one decimal place).
    (b)~Among $W\!=\!1$ episodes, percentages are the fraction of each primitive action type among all steps \emph{after} the world goal is first satisfied (episode-level action counts are pooled; bars sum to 100\%). NR-heavy models keep issuing navigation and pixel-interaction commands rather than closing, whereas low-$\Delta$ models concentrate on \texttt{report}.}
    \label{fig:belief_closure}
\end{figure*}

\paragraph{A single scalar cannot characterize closure reliability.}
Gemini-3.1-Pro and Claude-Sonnet-4 have similarly small gaps ($\Delta\!\approx\!1$--$4$\,pp; Table~\ref{tab:global_cross_section}), yet their closure regimes differ sharply (Figure~\ref{fig:failure_distribution}): Gemini combines high \textbf{W} (57.7\%) with moderate FR (24.6\%), whereas Claude-Sonnet-4 reaches much lower \textbf{W} (25.2\%) but reports aggressively (FR\,$=$\,69.9\%).
Claude-Sonnet-4's small gap is not evidence of reliable closure; it reflects frequent terminal commitment despite weak state support, where low \textbf{W} limits the number of achieved states available to expose missed conversions.
We therefore interpret closure behavior through the joint pattern over \textbf{W}, \textbf{B}, FR, NR, and IL rather than any single scalar.

\paragraph{When correct, closure follows task-state attainment quickly.}
Figure~\ref{fig:belief_closure}a shows a closure-latency proxy: the number of steps between first world-goal satisfaction and the terminal \texttt{report}, for episodes ending in a correct report.
Within this correctly closed subset, the lag ranges from 0.9 to 1.9\,steps across all 10 anchor models, indicating that successful closure is typically prompt once task-state attainment is reached.
When correct closure occurs at all, it is prompt; the main failures are missed or incorrect closure rather than slow eventual reporting. Exact counts are in Appendix~\ref{app:belief_lag}.

\paragraph{Most false-success commitments are unsupported by task progress.}
To characterize premature success claims at the step level, we examine the world-state progress recorded by the per-step evaluator at the moment of each false-success report.
Across all 10 anchor models, 65--88\% of false-success reports occur at exactly zero task progress: the agent has not navigated closer to the target, changed any task-relevant object state, or otherwise advanced the world toward the goal (Table~\ref{tab:belief_lag}).
These are unsupported terminal commitments issued in the absence of task-relevant world-state change.
The remaining false-success reports occur mostly at intermediate progress, typically after partial navigation or failed interaction; false-success commitments at high progress ($>$0.75) are rare or absent across all models.
The prevalence of zero-progress false-success reports is consistent with two interpretations: the model may genuinely misjudge task state, or it may treat the report as a default exit action regardless of task-state evidence.
VIGIL's behavioral interface cannot disambiguate these causes, but either mechanism yields closure failures that aggregate success would mask.

\paragraph{NR-heavy models drift after task-state attainment instead of closing.}
For the NR failure mode, we examine what agents do after the world condition is already satisfied (Figure~\ref{fig:belief_closure}b).
InternVL3.5-38B has 201 NR episodes among its 399 world-goal-met episodes; the majority of post-completion actions are navigation commands, with only a small fraction being \texttt{report}.
A similar pattern holds for Qwen3-VL-32B and MiMo-Embodied-7B.
By contrast, Gemini-3.1-Pro issues \texttt{report} as the dominant post-completion action, consistent with its low NR count (4 episodes) and small $\Delta$.
This suggests that NR is not primarily a timing artifact near the budget boundary; it is post-attainment drift in which achieved task state fails to trigger closure.

\subsection{Do Closure Failures Persist After Partial Execution Improvement?}
\label{sec:feedback}

The preceding analysis identifies distinct terminal-failure profiles under the no-feedback contract.
A natural question is whether these profiles are downstream consequences of recurrent execution traps (e.g., \texttt{path\_blocked}, \texttt{too\_far}) that consume the step budget before reliable closure can occur.
To test this, we add a minimal action-feedback intervention: two booleans after each action---\texttt{too\_far} (interaction beyond the proximity threshold) and \texttt{path\_blocked} (navigation obstructed).
These signals operationalize proprioceptive and low-level controller feedback in physical robots (e.g., out-of-reach manipulation and blocked motion).
Crucially, they expose execution outcomes only, not goal completion, task progress, or whether a report should be issued.
If closure failures persist, they are not fully explained by execution traps alone.
Table~\ref{tab:proprio_minimal} reports paired runs for all 10 anchor models.

\begin{table}[t]
\centering
\caption{%
\textbf{Diagnostic intervention: action feedback modeled on proprioceptive signals.}
Base\,=\,no-feedback contract; +FB\,=\,adds \texttt{too\_far} and \texttt{path\_blocked} boolean signals.
Event counts are mean per-episode values; $\Delta$W/$\Delta$B/$\Delta$FR/$\Delta$NR in pp.%
}
\label{tab:proprio_minimal}
\small
\setlength{\tabcolsep}{2.2pt}
\renewcommand{\arraystretch}{1.02}
\begin{tabular}{@{}l cc cc cc cc cc@{}}
\toprule
& \multicolumn{2}{c}{\textbf{too\_far}} & \multicolumn{2}{c}{\textbf{path\_blk}} & \multicolumn{2}{c}{\textbf{retry}} & \multicolumn{2}{c}{\textbf{Outcome}} & \multicolumn{2}{c@{}}{\textbf{Closure}} \\
\cmidrule(lr){2-3}\cmidrule(lr){4-5}\cmidrule(lr){6-7}\cmidrule(lr){8-9}\cmidrule(l){10-11}
\textbf{Model} & Base & +FB & Base & +FB & Base & +FB & $\boldsymbol{\Delta}$\textbf{W} & $\boldsymbol{\Delta}$\textbf{B} & $\boldsymbol{\Delta}$\textbf{FR} & $\boldsymbol{\Delta}$\textbf{NR} \\
\midrule
Gemini-3.1-Pro\rlap{$^\dagger$} & 1.29 & 0.73 & 2.76 & 1.12 & 0.78 & 0.16 & +13.9 & +13.0 & $-$9.2 & $-$4.6 \\
Doubao-Seed-1.8\rlap{$^\dagger$} & 1.09 & 0.50 & 4.40 & 2.28 & 1.01 & 0.11 & +12.1 & +12.6 & $-$4.6 & $-$8.3 \\
Qwen3.6-27B                     & 2.60 & 0.73 & 5.15 & 2.49 & 3.56 & 0.65 & +8.0 & +12.1 & +4.0 & $-$17.6 \\
Qwen3.5-27B                     & 3.19 & 0.75 & 4.15 & 2.24 & 2.37 & 0.16 & +11.3 & +12.1 & $-$0.1 & $-$15.8 \\
GPT-5.4\rlap{$^\dagger$}        & 0.18 & 0.27 & 2.22 & 1.48 & 0.01 & 0.01 & +4.5 & +2.3 & $-$3.6 & +3.6 \\
Claude-Sonnet-4\rlap{$^\dagger$} & 0.18 & 0.24 & 0.78 & 1.31 & 0.18 & 0.76 & +2.7 & +0.9 & $-$23.0 & +8.5 \\
Qwen3-VL-32B                    & 1.27 & 0.41 & 5.29 & 1.93 & 2.38 & 0.42 & +2.6 & +3.0 & +1.5 & $-$8.8 \\
Qwen3-VL-32B \texttt{(T)}       & 1.99 & 0.80 & 1.55 & 1.07 & 3.70 & 0.41 & +4.0 & +1.4 & +4.1 & $-$5.0 \\
InternVL3.5-38B                  & 0.86 & 0.50 & 11.94 & 7.38 & 1.52 & 0.01 & $-$0.6 & +1.5 & $-$4.0 & +0.7 \\
MiMo-Embodied-7B                & 2.07 & 0.67 & 4.24 & 2.43 & 4.24 & 0.39 & +3.0 & +0.0 & +0.6 & $-$7.1 \\
\bottomrule
\end{tabular}
\renewcommand{\arraystretch}{1.0}
\end{table}

\paragraph{Execution improvement does not propagate uniformly to closure.}
Nine of ten models reduce \texttt{path\_blocked} events under feedback, and most reduce \texttt{too\_far}, confirming that the signal is consumed.
However, Gemini-3.1-Pro, Doubao-Seed-1.8, Qwen3.5-27B---and partially Qwen3.6-27B---show large gains in \textbf{B} (up to $+$13\,pp) together with substantial gains in \textbf{W}.
For the first three, \textbf{W} rises by up to $+$14\,pp; Qwen3.6-27B attains $+$12.1\,pp in \textbf{B} but only $+$8.0\,pp in \textbf{W}.
This suggests that partial execution repair can unlock correct terminal commitment when closure behavior is already state-coupled.
We operationalize state-coupled closure as follows: under the baseline no-feedback contract, Gemini, Doubao, Qwen3.6, and Qwen3.5 combine moderate $\Delta$ (1.3--8.9\,pp) with aggregate FR far below GPT-5.4/Claude profiles (Gemini 24.6\%, Doubao 37.2\%, Qwen3.6 18.4\%, Qwen3.5 25.5\%; Figure~\ref{fig:failure_distribution}).
Their dominant bottleneck is reaching the goal state; feedback removes execution traps (reducing \texttt{path\_blocked} by 40--60\%), enabling more episodes to reach goal satisfaction---and because closure is already state-coupled, the additional achieved states are promptly converted into correct reports.
Qwen3.6-27B is informative: its $\Delta$B ($+$12.1\,pp) exceeds $\Delta$W ($+$8.0\,pp), indicating that improved execution disproportionately resolves missed closures over newly reached goals under this pairing.

\paragraph{NR-heavy and FR-heavy closure failures persist under partial execution repair.}
The remaining six models show a dissociation between execution improvement and terminal commitment, through three patterns.
\emph{FR-heavy persistence}: GPT-5.4 gains +4.5\,pp in \textbf{W} but only +2.3\,pp in \textbf{B}; its FR barely moves ($\Delta$FR\,$=$\,$-$3.6\,pp, remaining above 50\%).
\emph{FR$\to$NR conversion}: Claude-Sonnet-4 is the most extreme case---feedback increases mean steps from 6.8 to 10.8 (paired logs in Appendix~\ref{app:feedback_detail}) and FR drops by 23\,pp, but NR rises by 8.5\,pp, consistent with premature commitments being suppressed without being replaced by correct reports, yielding only +0.9\,pp in \textbf{B}.
\emph{NR-heavy persistence}: InternVL3.5-38B reduces \texttt{path\_blocked} from 11.94 to 7.38 per episode yet gains only +1.5\,pp in \textbf{B}; its NR remains unchanged ($\Delta$NR\,$=$\,+0.7\,pp), indicating that the closure bottleneck is not fully explained by navigation difficulty.
MiMo-Embodied-7B and the two Qwen3-VL-32B variants follow the same pattern: all gain modestly in \textbf{W} (+2.6 to +4.0\,pp) with minimal or zero \textbf{B} improvement, and FR either holds steady or \emph{increases}, suggesting that improved reachability can expose premature commitments rather than automatically produce correct reports (full deltas in Table~\ref{tab:proprio_minimal}).
These results show that improving execution is sometimes necessary, but not sufficient, for terminal commitment---the two dimensions require separate diagnosis and intervention.

\section{Conclusion and Limitations}
\label{sec:conclusion}

\nickname makes terminal commitment independently measurable by separating world-state completion from report correctness under a strict first-person, no-feedback contract.
Across 20 models on 1{,}000 frozen episodes, execution and terminal commitment separate systematically: models with comparable world-state completion differ by up to 19.7\,pp in benchmark success, and an action-feedback intervention modeled on proprioceptive signals improves \textbf{W} broadly but leaves closure failures intact for models whose terminal behavior is not state-coupled.
These findings confirm that today's embodied models can achieve task-relevant world states yet fail to convert them into correct terminal reports---and no prior evaluation makes this visible.

\paragraph{Limitations.}
All experiments run in AI2-THOR~\citep{kolve2017ai2thor} with ProcTHOR~\citep{deitke2022procthor}-generated houses under a single first-person, no-feedback contract; generalization to photorealistic simulators, physical robots, or alternative control interfaces has not been established.
The mandatory \texttt{report} protocol is itself a measurement instrument: it may elicit closure failures that a stop-only interface would leave latent, so results characterize behavior under this contract rather than model-optimal performance.

\paragraph{Broader impacts.}
Making premature and missed terminal commitments observable may help improve the reliability of embodied agents in settings where acting without confirming task completion can be costly.
At the same time, progress on semantic closure under a fixed reporting contract should not be mistaken for general embodied competence or calibrated self-knowledge outside that contract.

\bibliographystyle{unsrtnat}
\bibliography{refs}

@article{cheng2025embodiedeval,
  title = {{EmbodiedEval: Evaluate Multimodal LLMs as Embodied Agents}},
  author = {Cheng, Zhili and Tu, Yuge and Li, Ran and Dai, Shiqi and Hu, Jinyi and Hu, Shengding and Li, Jiahao and Shi, Yang and Yu, Tianyu and Chen, Weize and others},
  journal = {arXiv preprint arXiv:2501.11858},
  year = {2025}
}

@inproceedings{khanna2024goatbench,
  title = {{GOAT-Bench: A Benchmark for Multi-Modal Lifelong Navigation}},
  author = {Khanna, Mukul and Ramrakhya, Ram and Chhablani, Gunjan and Yenamandra, Sriram and Gervet, Theophile and Chang, Matthew and Kira, Zsolt and Chaplot, Devendra Singh and Batra, Dhruv and Mottaghi, Roozbeh},
  booktitle = {CVPR},
  year = {2024}
}

@inproceedings{li2024embodied,
    title={{Embodied Agent Interface: Benchmarking LLMs for Embodied Decision Making}},
    author={Li, Manling and Zhao, Shiyu and Wang, Qineng and Wang, Kangrui and Zhou, Yu and Srivastava, Sanjana and Gokmen, Cem and Lee, Tony and Li, Li Erran and Zhang, Ruohan and others},
    booktitle={NeurIPS},
    year={2024}
}

@inproceedings{peng2026nativeembodied,
  title = {{How Foundational Skills Influence VLM-based Embodied Agents: A Native Perspective}},
  author = {Peng, Bo and Bu, Pi and Pan, Keyu and Xu, Xinrun and Zhao, Yinxiu and Chen, Miao and Du, Yang and Li, Lin and Song, Jun and Xu, Tong},
  booktitle = {AAAI},
  year = {2026},
  note = {See also arXiv preprint arXiv:2602.20687.},
}

@inproceedings{shridhar2020alfred,
  title = {{ALFRED: A Benchmark for Interpreting Grounded Instructions for Everyday Tasks}},
  author = {Shridhar, Mohit and Thomason, Jesse and Gordon, Daniel and Bisk, Yonatan and Han, Winson and Mottaghi, Roozbeh and Zettlemoyer, Luke and Fox, Dieter},
  booktitle = {CVPR},
  year = {2020}
}

@inproceedings{padmakumar2022teach,
  title = {{TEACh: Task-Driven Embodied Agents That Chat}},
  author = {Padmakumar, Aishwarya and Thomason, Jesse and Shrivastava, Ayush and Lange, Patrick and Narayan-Chen, Anjali and Gella, Spandana and Piramuthu, Robinson and Tur, Gokhan and Hakkani-Tur, Dilek},
  booktitle = {AAAI},
  year = {2022}
}

@inproceedings{li2023behavior1k,
  title = {{BEHAVIOR-1K: A Benchmark for Embodied AI with 1,000 Everyday Activities and Realistic Simulation}},
  author = {Li, Chengshu and Zhang, Ruohan and Wong, Josiah and Gokmen, Cem and Srivastava, Sanjana and Martin-Martin, Roberto and Wang, Chen and Levine, Gabrael and Lingelbach, Michael and Sun, Jiankai and others},
  booktitle = {CoRL},
  year = {2023}
}

@inproceedings{shridhar2021alfworld,
  title = {{ALFWorld: Aligning Text and Embodied Environments for Interactive Learning}},
  author = {Shridhar, Mohit and Yuan, Xingdi and C{\^o}t{\'e}, Marc-Alexandre and Bisk, Yonatan and Trischler, Adam and Hausknecht, Matthew},
  booktitle = {ICLR},
  year = {2021}
}

@inproceedings{zheng2022vlmbench,
    title={{VLMbench: A Compositional Benchmark for Vision-and-Language Manipulation}},
    author={Kaizhi Zheng and Xiaotong Chen and Odest Jenkins and Xin Eric Wang},
    booktitle={NeurIPS},
    year={2022}
}

@inproceedings{choi2024lota,
  title={{LoTa-Bench: Benchmarking Language-oriented Task Planners for Embodied Agents}},
  author={Choi, Jae-Woo and Yoon, Youngwoo and Ong, Hyobin and Kim, Jaehong and Jang, Minsu},
  booktitle={ICLR},
  year={2024}
}

@inproceedings{kim2024realfred,
  title = {{ReALFRED: An Embodied Instruction Following Benchmark in Photo-Realistic Environments}},
  author = {Kim, Taewoong and Min, Cheolhong and Kim, Byeonghwi and Kim, Jinyeon and Jeung, Wonje and Choi, Jonghyun},
  booktitle = {ECCV},
  year = {2024}
}

@inproceedings{brohan2023rt2,
  title = {{RT-2: Vision-Language-Action Models Transfer Web Knowledge to Robotic Control}},
  author = {Zitkovich, Brianna and Yu, Tianhe and Xu, Sichun and Xu, Peng and Xiao, Ted and Xia, Fei and Wu, Jialin and Wohlhart, Paul and Welker, Stefan and Wahid, Ayzaan and Vuong, Quan and Vanhoucke, Vincent and Tran, Huong and Soricut, Radu and Singh, Anikait and Singh, Jaspiar and Sermanet, Pierre and Sanketi, Pannag R. and Salazar, Grecia and Ryoo, Michael S. and others},
  booktitle = {CoRL},
  year = {2023}
}

@inproceedings{chen2024spatialvlm,
  title = {{SpatialVLM: Endowing Vision-Language Models with Spatial Reasoning Capabilities}},
  author = {Chen, Boyuan and Xu, Zhuo and Kirmani, Sean and Driess, Danny and Florence, Pete and Ichter, Brian and Sadigh, Dorsa and Guibas, Leonidas and Xia, Fei},
  booktitle = {CVPR},
  year = {2024}
}

@inproceedings{zhang2026theory,
  title={{Theory of Space: Can Foundation Models Construct Spatial Beliefs through Active Exploration?}},
  author={Zhang, Pingyue and Huang, Zihan and Wang, Yue and Zhang, Jieyu and Xue, Letian and Wang, Zihan and Wang, Qineng and Chandrasegaran, Keshigeyan and Zhang, Ruohan and Choi, Yejin and others},
  booktitle={ICLR},
  year={2026},
  note = {See also arXiv preprint arXiv:2602.07055.},
}

@inproceedings{gao2025cubebench,
  title={{CubeBench: Diagnosing Interactive, Long-Horizon Spatial Reasoning under Partial Observations}},
  author={Gao, Huan-ang and Zhang, Zikang and Luo, Tianwei and Yang, Kaisen and Juan, Xinzhe and Qiu, Jiahao and Chen, Tianxing and He, Bingxiang and Zhao, Hao and Zhou, Hao and Liu, Shilong and Wang, Mengdi},
  booktitle={ICLR},
  year={2026}
}

@inproceedings{wang2025embodiedbench,
  title = {{EmbodiedBench: Comprehensive Benchmarking Multi-modal Large Language Models for Vision-Driven Embodied Agents}},
  author = {Yang, Rui and Chen, Hanyang and Zhang, Junyu and others},
  booktitle = {ICML},
  year = {2025},
  note = {See also arXiv preprint arXiv:2502.09560.},
}

@article{bai2025qwen3,
  title={{Qwen3-VL Technical Report}},
  author={Bai, Shuai and Cai, Yuxuan and Chen, Ruizhe and Chen, Keqin and Chen, Xionghui and Cheng, Zesen and Deng, Lianghao and Ding, Wei and Gao, Chang and Ge, Chunjiang and others},
  journal={arXiv preprint arXiv:2511.21631},
  year={2025}
}

@inproceedings{cheng2024spatialrgpt,
    title={{SpatialRGPT: Grounded Spatial Reasoning in Vision-Language Models}},
    author={Cheng, An-Chieh and Yin, Hongxu and Fu, Yang and Guo, Qiushan and Yang, Ruihan and Kautz, Jan and Wang, Xiaolong and Liu, Sifei},
    booktitle={NeurIPS},
    year={2024}
}

@inproceedings{omnispatial25,
  title   = {{OmniSpatial: Towards Comprehensive Spatial Reasoning Benchmark for Vision Language Models}},
  author  = {Mengdi Jia and Zekun Qi and Shaochen Zhang and Wenyao Zhang and Xinqiang Yu and Jiawei He and He Wang and Li Yi},
  booktitle = {ICLR},
  year = {2026},
  note = {See also arXiv preprint arXiv:2506.03135.},
}

@inproceedings{yang2025mmsi,
  title={{MMSI-Bench: A Benchmark for Multi-Image Spatial Intelligence}},
  author={Yang, Sihan and Xu, Runsen and Xie, Yiman and Yang, Sizhe and Li, Mo and Lin, Jingli and Zhu, Chenming and Chen, Xiaochen and Duan, Haodong and Yue, Xiangyu and Lin, Dahua and Wang, Tai and Pang, Jiangmiao},
  booktitle={ICLR},
  year={2025}
}

@inproceedings{wang2025spatial457,
  title     = {{Spatial457: A Diagnostic Benchmark for 6D Spatial Reasoning of Large Multimodal Models}},
  author    = {Wang, Xingrui and Ma, Wufei and Zhang, Tiezheng and de Melo, Celso M and Chen, Jieneng and Yuille, Alan},
  booktitle = {CVPR},
  year      = {2025}
}

@inproceedings{yuan2024robopoint,
  title={{RoboPoint: A Vision-Language Model for Spatial Affordance Prediction for Robotics}},
  author={Yuan, Wentao and Duan, Jiafei and Blukis, Valts and Pumacay, Wilbert and Krishna, Ranjay and Murali, Adithyavairavan and Mousavian, Arsalan and Fox, Dieter},
  booktitle={CoRL},
  year={2024}
}

@inproceedings{song2025robospatial,
  author    = {Song, Chan Hee and Blukis, Valts and Tremblay, Jonathan and Tyree, Stephen and Su, Yu and Birchfield, Stan},
  title     = {{RoboSpatial: Teaching Spatial Understanding to 2D and 3D Vision-Language Models for Robotics}},
  booktitle = {CVPR},
  year      = {2025}
}

@article{yang2025cambrians,
  title={{Cambrian-S: Towards Spatial Supersensing in Video}},
  author={Yang, Shusheng and Yang, Jihan and Huang, Pinzhi and Brown, Ellis and Yang, Zihao and Yu, Yue and Tong, Shengbang and Zheng, Zihan and Xu, Yifan and Wang, Muhan and Lu, Daohan and Fergus, Rob and LeCun, Yann and Fei-Fei, Li and Xie, Saining},
  journal={arXiv preprint arXiv:2511.04670},
  year={2025}
}

@inproceedings{zhou2025vlm4d,
    title={{VLM4D: Towards Spatiotemporal Awareness in Vision Language Models}},
    author={Zhou, Shijie and Vilesov, Alexander and He, Xuehai and Wan, Ziyu and Zhang, Shuwang and Nagachandra, Aditya and Chang, Di and Chen, Dongdong and Wang, Eric Xin and Kadambi, Achuta},
    booktitle={ICCV},
    year={2025}
}

@inproceedings{yang2024think,
    title={{Thinking in Space: How Multimodal Large Language Models See, Remember and Recall Spaces}},
    author={Yang, Jihan and Yang, Shusheng and Gupta, Anjali W. and Han, Rilyn and Fei-Fei, Li and Xie, Saining},
    year={2024},
    booktitle={CVPR},
}

@inproceedings{yin2025mental,
      title={{Spatial Mental Modeling from Limited Views}}, 
      author={Qineng Wang and Baiqiao Yin and Pingyue Zhang and Jianshu Zhang and Kangrui Wang and Zihan Wang and Jieyu Zhang and Keshigeyan Chandrasegaran and Han Liu and Ranjay Krishna and Saining Xie and Manling Li and Jiajun Wu and Li Fei-Fei},
      year={2025},
      booktitle={ICLR}
}

@inproceedings{ren2023knowno,
  title={{Robots That Ask For Help: Uncertainty Alignment for Large Language Model Planners}},
  author={Ren, Allen Z. and Dixit, Anushri and Bodrova, Alexandra and Singh, Sumeet and Tu, Stephen and Brown, Noah and Xu, Peng and Takayama, Leila and Xia, Fei and Varley, Jake and Xu, Zhenjia and Sadigh, Dorsa and Zeng, Andy and Majumdar, Anirudha},
  booktitle={CoRL},
  year={2023}
}

@inproceedings{huang2022innermonologue,
  title={{Inner Monologue: Embodied Reasoning through Planning with Language Models}},
  author={Huang, Wenlong and Xia, Fei and Xiao, Ted and Chan, Harris and Liang, Jacky and Florence, Pete and Zeng, Andy and Tompson, Jonathan and Mordatch, Igor and Chebotar, Yevgen and Sermanet, Pierre and Brown, Noah and Jackson, Tomas and Luu, Linda and Levine, Sergey and Hausman, Karol and Ichter, Brian},
  booktitle={CoRL},
  year={2022}
}

@article{kadavath2022language,
  title={{Language Models (Mostly) Know What They Know}},
  author={Kadavath, Saurav and Conerly, Tom and Askell, Amanda and Henighan, Tom and Drain, Dawn and Perez, Ethan and Schiefer, Nicholas and Hatfield-Dodds, Zac and DasSarma, Nova and Tran-Vu, Eli and others},
  journal={arXiv preprint arXiv:2207.05221},
  year={2022}
}

@article{kolve2017ai2thor,
  title={{AI2-THOR: An Interactive 3D Environment for Visual AI}},
  author={Kolve, Eric and Mottaghi, Roozbeh and Han, Winson and VanderBilt, Eli and Weihs, Luca and Herrasti, Alvaro and Gordon, Daniel and Zhu, Yuke and Gupta, Abhinav and Farhadi, Ali},
  journal={arXiv preprint arXiv:1712.05474},
  year={2017}
}

@inproceedings{deitke2022procthor,
  title={{ProcTHOR: Large-Scale Embodied AI Using Procedural Generation}},
  author={Deitke, Matt and VanderBilt, Eli and Herrasti, Alvaro and Weihs, Luca and Ehsani, Kiana and Salvador, Jordi and Han, Winson and Kolve, Eric and Kembhavi, Aniruddha and Mottaghi, Roozbeh},
  booktitle={NeurIPS},
  year={2022}
}

@article{wang2025internvl35,
  title={{InternVL3.5}: Advancing Open-Source Multimodal Models in Versatility, Reasoning, and Efficiency},
  author={Wang, Weiyun and Gao, Zhangwei and Gu, Lixin and Chen, Zhe and others},
  journal={arXiv preprint arXiv:2508.18265},
  year={2025}
}

@article{xiaomi2025mimo,
  title={{MiMo-Embodied: X-Embodied Foundation Model Technical Report}},
  author={Hao, Xiaoshuai and Zhou, Lei and Huang, Zhijian and Hou, Zhiwen and Tang, Yingbo and Zhang, Lingfeng and Li, Guang and Lu, Zheng and others},
  journal={arXiv preprint arXiv:2511.16518},
  year={2025}
}

@article{kimiteam2025kimivl,
  title={{Kimi-VL Technical Report}},
  author={{Kimi Team} and Du, Angang and others},
  journal={arXiv preprint arXiv:2504.07491},
  year={2025}
}

@article{guo2025seed15vl,
  title={{Seed1.5-VL Technical Report}},
  author={Guo, Dong and Wu, Faming and Zhu, Feida and others},
  journal={arXiv preprint arXiv:2505.07062},
  year={2025}
}

@article{ji2025robobrain,
  title={{RoboBrain: A Unified Brain Model for Robotic Manipulation from Abstract to Concrete}},
  author={Ji, Yuheng and Tan, Huajie and Shi, Jiayu and Hao, Xiaoshuai and Zhang, Yuan and Zhang, Hengyuan and Wang, Pengwei and Zhao, Mengdi and Mu, Yao and An, Pengju and others},
  journal={arXiv preprint arXiv:2502.21257},
  year={2025}
}

@article{dang2026rynnbrain,
  title={{RynnBrain: Open Embodied Foundation Models}},
  author={Dang, Ronghao and Guo, Jiayan and Hou, Bohan and Leng, Sicong and Li, Kehan and Li, Xin and Liu, Jiangping and Mao, Yunxuan and Wang, Zhikai and Yuan, Yuqian and Zhu, Minghao and Lin, Xiao and Bai, Yang and Jiang, Qian and Zhao, Yaxi and Zeng, Minghua and Gao, Junlong and Jiang, Yuming and Cen, Jun and Huang, Siteng and Wang, Liuyi and Zhang, Wenqiao and Liu, Chengju and Yang, Jianfei and Lu, Shijian and Zhao, Deli},
  journal={arXiv preprint arXiv:2602.14979},
  year={2026}
}

@article{yang2025qwen3,
  title={{Qwen3 Technical Report}},
  author={Yang, An and Li, Anfeng and Yang, Baosong and Zhang, Beichen and others},
  journal={arXiv preprint arXiv:2505.09388},
  year={2025}
}

@misc{qwenteam2026qwen35,
  title = {{Qwen3.5}: Towards Native Multimodal Agents},
  author = {{Qwen Team}},
  year = {2026},
  month = feb,
  howpublished = {\url{https://qwen.ai/blog?id=qwen3.5}},
}

@misc{qwenteam2026qwen3627b,
  title = {{Qwen3.6-27B}: Flagship-Level Coding in a {27B} Dense Model},
  author = {{Qwen Team}},
  year = {2026},
  month = apr,
  howpublished = {\url{https://qwen.ai/blog?id=qwen3.6-27b}},
}

@misc{deepmind2026gemini31pro,
  title = {{Gemini} 3.1 {Pro} Model Card},
  author = {{Google DeepMind}},
  year = {2026},
  howpublished = {\url{https://deepmind.google/models/model-cards/gemini-3-1-pro/}},
}

@misc{openai2026gpt54,
  title = {Introducing {GPT-5.4}},
  author = {{OpenAI}},
  year = {2026},
  howpublished = {\url{https://openai.com/index/introducing-gpt-5-4/}},
}

@misc{anthropic2025claude4,
  title = {Introducing {Claude} 4},
  author = {{Anthropic}},
  year = {2025},
  month = may,
  howpublished = {\url{https://www.anthropic.com/news/claude-4}},
}

\newpage
\appendix

\section{System Prompt Specification}
\label{app:prompt_spec}

The system prompt is assembled programmatically from four blocks in fixed order.
No privileged simulator state crosses the agent--evaluator boundary.
We reproduce each block verbatim below; each benchmark run stores a SHA-256 hash of the rendered prompt alongside the profile name and prompt-policy version, enabling bit-exact reproducibility audits.

\begin{tcolorbox}[promptbox, title={Block 1: Task}]
You are an embodied agent in a simulated 3D environment with an egocentric (first-person) view.\\[3pt]
Each step you only see local observations; plan and act under partial observability.\\[3pt]
Ground decisions in the task, the current observation, recent observation history when available, and the allowed actions schema.\\[3pt]
Actions are embodied attempts in the scene, not symbolic shortcuts.\\[3pt]
Use only public evidence from this episode. Do not assume hidden simulator state or hidden success signals.\\[3pt]
For state-recognition tasks, report the observed state label (for example: on/off or open/closed) rather than mapping the state to task success/fail.\\[3pt]
Mere visibility, proximity, or an attempted action is not enough; do not report success from visual plausibility alone.\\[3pt]
Do not claim completion from visual plausibility alone when the task requires changing world state.\\[3pt]
Task: \textrm{\textit{<TASK\_INSTRUCTION>}}
\end{tcolorbox}

\begin{tcolorbox}[promptbox, title={Block 2: Action Schema}]
\#\# Allowed actions\
- navigate \textrm{|} look \textrm{|} interact\_pixel \textrm{|} report\\[3pt]
\textrm{\textit{Each skill is serialized from the action registry with its required/optional arguments, type constraints, and usage notes. The full specification matches Tables~\ref{tab:action_specs}--\ref{tab:report_statuses} (Appendix~\ref{app:action_space}) verbatim.}}
\end{tcolorbox}

\begin{tcolorbox}[promptbox, title={Block 3: Grounding Rules}]
\#\# Grounding Rules\
- Use only the frame whose {\char96}role{\char96} is {\char96}current{\char96} for visibility judgments and interact\_pixel targeting.\
- interact\_pixel acts on the object currently under the specified location in the current RGB frame.\
- Do not assume hidden success state; rely on subsequent public observations and allowed feedback fields from this episode.
\end{tcolorbox}

\noindent
For models using normalized coordinates, an additional rule is prepended:
\texttt{For interact\_pixel, output x and y in normalized\_1000 coordinates: integers in [0,\,1000], where (0,0) is the top-left and (1000,1000) is the bottom-right corner.}

\begin{tcolorbox}[promptbox, title={Block 4: Output Format}]
\#\# Output format\
Return exactly one JSON object and nothing else.\
Required keys: {\char96}skill\_name{\char96}, {\char96}arguments{\char96}.\
Optional keys: {\char96}thought{\char96}, {\char96}cognitive\_state{\char96}.\
Use the exact action names from the allowed-actions schema above; do not invent aliases or synonyms.\
Do not output prose, markdown, code fences, or multiple JSON objects.\
If you include {\char96}thought{\char96}, keep it to one short sentence.
\end{tcolorbox}

\section{Action Space and Termination Contract}
\label{app:action_space}

\subsection{Skill Definitions}

The action space consists of four skills: two for locomotion (\texttt{navigate}, \texttt{look}), one for object interaction (\texttt{interact\_pixel}), and one for terminal self-report (\texttt{report}).
Table~\ref{tab:action_specs} provides the complete specification.

\begin{table*}[h]
\centering
\caption{%
\textbf{Complete action-space specification.}
\textbf{R} denotes required; \textbf{C} denotes conditionally required (required unless intent is \texttt{drop}).%
}
\label{tab:action_specs}
\footnotesize
\setlength{\tabcolsep}{3.5pt}
\renewcommand{\arraystretch}{1.15}
\begin{tabularx}{\textwidth}{@{}l l l >{\raggedright\arraybackslash}X l@{}}
\toprule
\textbf{Skill} & \textbf{Argument} & \textbf{Type} & \textbf{Description} & \textbf{Req.} \\
\midrule
\multirow{2}{*}{\texttt{navigate}}
  & \texttt{mode}      & enum   & \texttt{forward}, \texttt{backward}, \texttt{turn\_left}, \texttt{turn\_right} & R \\
  & \texttt{magnitude} & number & Step count (fwd/bwd) or rotation in degrees (turns) & R \\
\midrule
\multirow{2}{*}{\texttt{look}}
  & \texttt{direction} & enum   & \texttt{up}, \texttt{down} & R \\
  & \texttt{magnitude} & number & Pitch change in degrees & R \\
\midrule
\multirow{3}{*}{\texttt{interact\_pixel}}
  & \texttt{intent} & enum  & One of eight canonical intents (Table~\ref{tab:intents}) & R \\
  & \texttt{x}      & int   & Horizontal pixel coordinate in the current RGB frame & C \\
  & \texttt{y}      & int   & Vertical pixel coordinate in the current RGB frame & C \\
\midrule
\multirow{2}{*}{\texttt{report}}
  & \texttt{status}  & enum   & Terminal status (Table~\ref{tab:report_statuses}) & R \\
  & \texttt{summary} & string & Brief justification for the chosen status & R \\
\bottomrule
\end{tabularx}
\end{table*}

\subsection{Canonical Interaction Intents}

Table~\ref{tab:intents} lists the eight canonical intents accepted by \texttt{interact\_pixel}.
Common aliases (e.g., \texttt{open} $\rightarrow$ \texttt{open\_access}, \texttt{toggle\_on} $\rightarrow$ \texttt{activate}) are normalized at runtime; models may emit either form.

\begin{table}[h]
\centering
\caption{\textbf{Canonical intents} for \texttt{interact\_pixel}.}
\label{tab:intents}
\footnotesize
\setlength{\tabcolsep}{4pt}
\renewcommand{\arraystretch}{1.12}
\begin{tabularx}{\columnwidth}{@{}l >{\raggedright\arraybackslash}X l@{}}
\toprule
\textbf{Intent} & \textbf{Semantics} & \textbf{Requires $(x,y)$} \\
\midrule
\texttt{ground}        & Point/click/locate without changing world state & Yes \\
\texttt{open\_access}  & Open a container or access point              & Yes \\
\texttt{close\_access} & Close a container or access point              & Yes \\
\texttt{activate}      & Turn on / toggle on                            & Yes \\
\texttt{deactivate}    & Turn off / toggle off                          & Yes \\
\texttt{pick}          & Pick up an object                              & Yes \\
\texttt{place}         & Place a held object at target                  & Yes \\
\texttt{drop}          & Drop the currently held object                 & No  \\
\bottomrule
\end{tabularx}
\end{table}

\subsection{Report Statuses}

\begin{table}[h]
\centering
\caption{\textbf{Terminal report statuses} accepted by \texttt{report}.}
\label{tab:report_statuses}
\footnotesize
\setlength{\tabcolsep}{4pt}
\renewcommand{\arraystretch}{1.12}
\begin{tabularx}{\columnwidth}{@{}l >{\raggedright\arraybackslash}X@{}}
\toprule
\textbf{Status} & \textbf{Semantics} \\
\midrule
\texttt{success} & Agent claims the task goal has been achieved \\
\texttt{fail}    & Agent concedes it cannot complete the task \\
\texttt{unsafe}  & Agent flags a safety concern \\
\texttt{invalid} & Agent flags the task as invalid or unsolvable \\
\texttt{on}      & State-recognition report: observed state is \emph{on} \\
\texttt{off}     & State-recognition report: observed state is \emph{off} \\
\texttt{open}    & State-recognition report: observed state is \emph{open} \\
\texttt{closed}  & State-recognition report: observed state is \emph{closed} \\
\bottomrule
\end{tabularx}
\end{table}
In the current frozen pack, \texttt{unsafe} and \texttt{invalid} are interface-level fallback statuses, not task-specific target labels.

\subsection{Step Budgets and Termination}

Each task family enforces a fixed step budget and an invalid-action limit (see Table~\ref{tab:family_specs} in Appendix~\ref{app:task_families} for per-family values).
Exceeding the step budget terminates the episode as \texttt{fail\_no\_report}; exceeding the invalid-action limit terminates as \texttt{invalid\_action\_limit\_exceeded}.

\subsection{Termination Contract}

The \texttt{report} skill is the \emph{sole} agent-initiated termination mechanism.
The agent must invoke it with a status and justification to end the episode; the benchmark never issues a report automatically.
Three terminal conditions exist:
\begin{enumerate}
    \item \textbf{Agent report:} the agent invokes \texttt{report}; the status is matched against hidden world state.
    \item \textbf{Budget exhaustion:} the step budget is reached without a report; scored as \texttt{fail\_no\_report}.
    \item \textbf{Invalid-action limit:} cumulative invalid actions (protocol failures + malformed actions) exceed the family-specific limit; scored as \texttt{invalid\_action\_limit\_exceeded}.
\end{enumerate}

\section{Scoring Details}
\label{app:scoring_details}

\subsection{Dual-Metric Evaluation}

Each episode is evaluated under two success metrics simultaneously:
\begin{itemize}
    \item \textbf{Semantic} (primary): tolerant of minor imprecision in object placement or state matching.
    \item \textbf{Strict}: exact simulator predicate check with no tolerance.
\end{itemize}
The main paper reports aggregate \textbf{W}, \textbf{B}, and $\Delta = W - B$ under the semantic world predicate paired with strict terminal \texttt{report} matching (see \S\ref{sec:termination}).
The strict world metric is retained as a validation variant to expose whether apparent world-state completion would rest on evaluation leniency.

\subsection{Reported Closure-Failure Labels}
\label{app:failure_taxonomy}

The main analysis reports three closure-failure labels, summarized in Table~\ref{tab:failure_taxonomy}.
They are computed post-hoc from the episode settlement record and are used consistently in Figure~\ref{fig:failure_distribution}, Table~\ref{tab:proprio_minimal}, and the extended model panels.

\begin{table}[h]
\centering
\caption{%
\textbf{Reported closure-failure labels.}
These are the diagnostic labels reported in the main figures and tables.
FR and NR separate incorrect terminal content from missing terminal commitment; IL captures forced termination due to repeated invalid actions.%
}
\label{tab:failure_taxonomy}
\footnotesize
\setlength{\tabcolsep}{3pt}
\renewcommand{\arraystretch}{1.15}
\begin{tabularx}{\columnwidth}{@{}l >{\raggedright\arraybackslash}X@{}}
\toprule
\textbf{Label} & \textbf{Definition} \\
\midrule
\textbf{FR} & False report: the agent submits a \texttt{report} whose status does not match the hidden world state or expected state label. This includes premature success claims and incorrect categorical state reports. \\[3pt]
\textbf{NR} & No report: the step budget expires without any \texttt{report}. When $W\!=\!1$, this is the post-attainment drift case analyzed in \S\ref{sec:terminal_commitment}. \\[3pt]
\textbf{IL} & Invalid-action-limit termination: repeated malformed or invalid actions exceed the family-specific invalid-action limit, forcing episode termination before reliable closure can be assessed. \\
\bottomrule
\end{tabularx}
\end{table}
This decomposition is intended to separate protocol-compliance failures from task-state-judgment failures: IL primarily captures malformed-action and interface-alignment breakdowns, while FR/NR characterize closure behavior on trajectories that remain protocol-valid until settlement.

\subsection{Deterministic Report-Matching Rules}
\label{app:match_rules}

The function $\texttt{match}(\texttt{report}_T, s_T)$ referenced in \S\ref{sec:termination} operates in two modes, determined by the episode's success-condition type.
No family-specific branching exists in the match logic itself; family-specific behavior arises only from how the world-success predicate $\mathcal{G}_{\mathrm{sem}}$ is evaluated.

\paragraph{Goal-completion mode (PG, DA, VS, AI, SI, SM, CR).}
The episode success condition checks a world-state predicate (e.g., \texttt{agent\_near\_target}, \texttt{object\_state}, \texttt{object\_held}).
The report status is first normalized to one of eight canonical values (\texttt{success}, \texttt{fail}, \texttt{unsafe}, \texttt{invalid}, \texttt{on}, \texttt{off}, \texttt{open}, \texttt{closed}); any other value is treated as invalid.
The match rule is:
\[
\texttt{match} = \bigl(\texttt{status} = \texttt{success} \;\land\; W\!=\!1\bigr) \;\lor\; \bigl(\texttt{status} \in \{\texttt{fail}, \texttt{unsafe}, \texttt{invalid}\} \;\land\; W\!=\!0\bigr).
\]
A correct \texttt{fail} report when $W\!=\!0$ therefore counts as a match (``honest fail''), and a \texttt{success} report when $W\!=\!0$ is a false-success report.
Under this goal-completion rule, \texttt{unsafe} and \texttt{invalid} are handled as negative reports in the same matching set as \texttt{fail}.

\paragraph{State-verification mode (SV).}
SV episodes use a \texttt{report\_status} success condition with an explicit expected label derived from the hidden simulator state via:
\[
\texttt{is\_toggled} \mapsto \texttt{on}/\texttt{off}, \qquad \texttt{is\_open} \mapsto \texttt{open}/\texttt{closed}.
\]
For benchmark success, the target state must be publicly observable at closure ($W\!=\!1$) and the report must equal this expected categorical label.
If the state is not publicly observable at closure, the episode is not a benchmark success regardless of report content.
The match rule is exact string equality between the normalized report status and the expected label:
\[
\texttt{match} = \bigl(\texttt{status} = \texttt{expected\_label}\bigr).
\]
A stop-only protocol cannot represent this judgment because the correct terminal output is categorical, not binary.

\section{Task Family Specifications}
\label{app:task_families}

The eight task families are organized into a \textbf{diagnostic tier} (PG, DA, VS, SV) that isolates atomic competencies and a \textbf{compositional tier} (AI, SI, SM, CR) that chains them under increasing partial-observability demand.
Table~\ref{tab:family_specs} summarizes the key parameters; detailed descriptions follow.

\begin{table*}[h]
\centering
\caption{%
\textbf{Task family specification summary.}
``Vis.''\ indicates whether the target is visible at episode start.
``Steps'' and ``Inv.'' are the episode step budget and invalid-action limit.
``Skills'' lists the expected skill repertoire; exact initial poses and distances vary by authored episode and are simulator-validated during pack construction.%
}
\label{tab:family_specs}
\footnotesize
\setlength{\tabcolsep}{3pt}
\renewcommand{\arraystretch}{1.15}
\begin{tabularx}{\textwidth}{@{}l l c c c >{\raggedright\arraybackslash}X@{}}
\toprule
\textbf{Family} & \textbf{Layer} & \textbf{Vis.} & \textbf{Steps} & \textbf{Inv.} & \textbf{Expected Skills} \\
\midrule
PG Pixel Grounding      & Diagnostic    & \checkmark &  5 &  3 & \texttt{interact\_pixel}(ground), \texttt{report} \\
DA Distance Approach     & Diagnostic    & \checkmark & 12 &  4 & \texttt{navigate}, \texttt{report} \\
VS View Search           & Diagnostic    & $\times$   & 20 &  6 & \texttt{navigate}, \texttt{look}, \texttt{report} \\
SV State Verification    & Diagnostic    & \checkmark &  5 &  2 & \texttt{look}, \texttt{report} \\
\midrule
AI Approach and Interact & Compositional & \checkmark & 25 &  8 & \texttt{navigate}, \texttt{interact\_pixel}, \texttt{report} \\
SI Search and Interact   & Compositional & $\times$   & 35 & 10 & \texttt{navigate}, \texttt{look}, \texttt{interact\_pixel}, \texttt{report} \\
SM Sequential Manipulation & Compositional & \checkmark & 30 & 10 & \texttt{navigate}, \texttt{interact\_pixel}, \texttt{report} \\
CR Constraint Resolving  & Compositional & \checkmark & 40 & 12 & \texttt{navigate}, \texttt{interact\_pixel}, \texttt{report} \\
\bottomrule
\end{tabularx}
\end{table*}

\paragraph{PG: Pixel Grounding.}
The target object is visible and reachable at episode start; the agent must click (ground) it via \texttt{interact\_pixel}(ground), then submit a \texttt{report}.
This isolates visual grounding---the ability to map a natural-language reference to a pixel coordinate in the egocentric frame.
Success criterion: \texttt{target\_grounded} (the grounding click falls on the correct object instance).
Difficulty varies by target object size: large (e.g., fridge, television), medium (e.g., microwave, cabinet), or small (e.g., egg, pen, watch).

\paragraph{DA: Distance Approach.}
The target is visible but initially outside the success threshold; the agent must navigate close enough, then report.
This isolates spatial navigation under egocentric observation---coarse depth estimation and obstacle avoidance without any interaction.
Success criterion: \texttt{agent\_near\_target} (Euclidean distance $< 1.5$\,m).

\paragraph{VS: View Search.}
The target is \emph{not visible} at episode start.
The agent must change viewpoint through \texttt{navigate} and/or \texttt{look} actions until the target becomes visible, then report.
This isolates active search under partial observability---the ability to systematically explore the environment.
Success criterion: \texttt{object\_state}(visible=True), latched once achieved.

\paragraph{SV: State Verification.}
The agent starts with a visible object whose state (on/off or open/closed) must be identified and reported.
This isolates perceptual state recognition---the ability to observe and classify the current state of an object.
Success criterion: \texttt{report\_status} matching the oracle state label.

\paragraph{AI: Approach and Interact.}
The target is visible at episode start.
The agent must approach as needed, perform a single pixel-grounded interaction (activate, deactivate, open, close, or pick), then report.
This composes DA (approach) and PG-level grounding with a state-changing interaction.
Success criterion: \texttt{object\_state} or \texttt{object\_held} depending on intent.

\paragraph{SI: Search and Interact.}
The target is not visible at episode start.
The agent must find the target, approach as needed, perform the required pixel-grounded interaction, and report.
This is the full perception--navigation--interaction pipeline under maximal partial observability.
Success criterion: same as AI, conditioned on prior search.

\paragraph{SM: Sequential Manipulation.}
The instruction requires an ordered manipulation chain, such as opening a container, picking an object, and placing it at a destination.
Four functional templates exist: \texttt{reveal\_pick} (open container, pick hidden object), \texttt{put\_into} (place object in receptacle), \texttt{rearrange} (move object between receptacles), and \texttt{open\_pick\_place} (open, pick, place elsewhere).
Success criterion: \texttt{object\_held} or \texttt{object\_at\_receptacle}.

\paragraph{CR: Constraint Resolving.}
The target is visible at episode start, but a physical constraint blocks the direct completion path.
The agent must resolve the constraint (e.g., move or interact with an obstacle), navigate as needed, perform the required interaction, and report.
This tests planning under physical constraints---the agent must reason about path feasibility before completing the interaction.
Success criterion: same as AI, with the additional requirement that the constraint is resolved.

\section{Benchmark Episode Details}
\label{app:benchmark_details}

\subsection{Episode Generation Pipeline}

Episodes are constructed through a three-stage pipeline:
\begin{enumerate}
    \item \textbf{LLM proposal}: a language model drafts candidate tasks conditioned on scene inventories and family-specific constraints (target visibility, start-pose requirements, available intents, object categories).
    \item \textbf{Simulator validation}: each proposal is instantiated in AI2-THOR and validated for object existence, state accessibility, agent reachability, and success-condition solvability.  The validation engine checks episode-contract integrity including agent initialization, scene setup consistency, success-spec type validity, and family-specific rules (e.g., SM requires pre-conditions and multi-object bindings; CR requires blocked-path preconditions).
    \item \textbf{Human review}: a human auditor reviews a stratified sample for ambiguity, instruction quality, and difficulty calibration.  Manually approved episodes receive priority in pack assembly.
\end{enumerate}

\subsection{Pack Composition}

The evaluation uses pack \texttt{mixed\_mainline\_manual\_balanced\_1000}, containing 1{,}000 episodes with exactly 125 per task family.
These episodes are selected from a larger authored pool spanning diverse AI2-THOR environments, including ProcTHOR-generated scenes.

\begin{table}[h]
\centering
\caption{%
\textbf{Episode distribution} in the 1{,}000-episode evaluation pack.%
}
\label{tab:episode_distribution}
\footnotesize
\setlength{\tabcolsep}{4pt}
\renewcommand{\arraystretch}{1.12}
\begin{tabularx}{\columnwidth}{@{}l >{\centering\arraybackslash}X >{\centering\arraybackslash}X@{}}
\toprule
\textbf{Family} & \textbf{Episodes} & \textbf{Max Steps} \\
\midrule
PG Pixel Grounding        & 125 &  5 \\
DA Distance Approach       & 125 & 12 \\
VS View Search             & 125 & 20 \\
SV State Verification      & 125 &  5 \\
\midrule
AI Approach and Interact   & 125 & 25 \\
SI Search and Interact     & 125 & 35 \\
SM Sequential Manipulation & 125 & 30 \\
CR Constraint Resolving    & 125 & 40 \\
\midrule
\textbf{Total}             & \textbf{1{,}000} & \\
\bottomrule
\end{tabularx}
\end{table}

\subsection{Simulator Configuration}

All episodes run in AI2-THOR (ProcTHOR scenes) with fixed rendering parameters:
resolution $640 \times 480$, field of view $90^{\circ}$, visibility distance $6.0$\,m, interaction distance limit $1.5$\,m, and instance segmentation enabled for grounding evaluation.
Segmentation is used only by the evaluator for offline grounding and settlement; it is never exposed to the agent, which receives only RGB frames under the default contract.
The agent's initial position, rotation, and camera horizon are specified per episode and validated against the scene setup.

\section{Primary 20-Model Comparison Panel}
\label{app:full_panel}

Table~\ref{tab:full_panel} reports per-family \textbf{W}/\textbf{B} for the primary 20-model panel used for cross-model comparison.
The panel includes models that reliably operate the native-control interface and are not dominated by invalid-action-limit failures or zero benchmark success; additional low-compliance, zero-\textbf{B}, and supplementary size/serving-mode variants are reported separately in Table~\ref{tab:extended_panel}.
The main text (Table~\ref{tab:global_cross_section}) uses 10 anchor models; this table adds 10 additional open-weight models spanning MoE variants, smaller checkpoints, and embodied-tuned systems, including Kimi-VL~\citep{kimiteam2025kimivl}, RoboBrain2.5-4B~\citep{ji2025robobrain}, MiMo-Embodied-7B~\citep{xiaomi2025mimo}, rynnbrain-8B~\citep{dang2026rynnbrain}, the Qwen3-VL and Qwen3.x families~\citep{bai2025qwen3,yang2025qwen3,qwenteam2026qwen35,qwenteam2026qwen3627b}, InternVL3.5~\citep{wang2025internvl35}, and closed-source APIs Gemini~3.1~Pro~\citep{deepmind2026gemini31pro}, Doubao-Seed-1.8~\citep{guo2025seed15vl}, GPT-5.4~\citep{openai2026gpt54}, and Claude-Sonnet-4~\citep{anthropic2025claude4}.
In that 10-anchor set, Qwen3-VL-32B Instruct and Qwen3-VL-32B \texttt{(T)} Thinking are counted as two distinct anchor models (\texttt{(T)}\,=\,public Thinking serving mode); Qwen3.6-27B and Qwen3.5-27B are served with thinking enabled (one anchor row each); every other checkpoint appears once.
Suffixes such as \texttt{A3B}/\texttt{A17B}/\texttt{A22B} denote \emph{activated} expert parameters in MoE-style checkpoints.

\begin{table*}[h]
\centering
\caption{%
\textbf{Primary 20-model comparison panel under native control.}
Each cell reports \textbf{W}/\textbf{B} (\%).
Diagnostic: PG\,=\,pixel grounding, DA\,=\,distance approach, VS\,=\,view search, SV\,=\,state verification.
Compositional: AI\,=\,approach-and-interact, SI\,=\,search-and-interact, SM\,=\,sequential manipulation, CR\,=\,constraint resolving.
Models are grouped by category and sorted by aggregate \textbf{B} within each group.%
}
\label{tab:full_panel}
\scriptsize
\setlength{\tabcolsep}{3.0pt}
\renewcommand{\arraystretch}{1.18}
\begin{tabularx}{\textwidth}{@{}>{\raggedright\arraybackslash}p{2.75cm}
  *{9}{>{\centering\arraybackslash}X}@{}}
\toprule
\multirow{2}{*}[-0.3ex]{\textbf{Model}}
  & \textbf{All}
  & \multicolumn{4}{c}{\textbf{Diagnostic Probes}}
  & \multicolumn{4}{c@{}}{\textbf{Compositional Tasks}} \\
\cmidrule(lr){2-2}\cmidrule(lr){3-6}\cmidrule(l){7-10}
  & \textbf{W/B} & \textbf{PG} & \textbf{DA} & \textbf{VS} & \textbf{SV}
  & \textbf{AI} & \textbf{SI} & \textbf{SM} & \textbf{CR} \\
\midrule
\rowcolor{EdgeBg}
\multicolumn{10}{@{}l}{\textit{Embodied / robotics-tuned}} \\
MiMo-Embodied-7B & 29.5/21.3 & 52.0/34.4 & 37.6/24.0 & 31.2/26.4 & 95.2/74.4 & 14.4/7.2 & 0.8/0.8 & 3.2/1.6 & 1.6/1.6 \\
RoboBrain2.5-4B & 27.4/14.6 & 78.4/78.4 & 19.2/14.4 & 13.6/12.0 & 94.4/0.0 & 8.0/8.0 & 0.0/0.0 & 4.0/2.4 & 1.6/1.6 \\
rynnbrain-8B & 24.1/17.3 & 17.6/11.2 & 50.4/38.4 & 4.8/1.6 & 90.4/60.0 & 20.8/18.4 & 0.8/0.8 & 6.4/6.4 & 1.6/1.6 \\
\addlinespace[0.2ex]
\rowcolor{MidBg}
\multicolumn{10}{@{}l}{\textit{Open general and reasoning VLMs}} \\
Kimi-VL-A3B \texttt{(T)} & 17.3/8.8 & 10.4/9.6 & 26.4/3.2 & 3.2/2.4 & 95.2/53.6 & 0.8/0.8 & 0.0/0.0 & 1.6/0.0 & 0.8/0.8 \\
Qwen3-VL-4B & 30.7/25.6 & 59.2/56.8 & 48.0/42.4 & 23.2/18.4 & 74.4/47.2 & 29.6/28.8 & 0.8/0.8 & 8.0/8.0 & 2.4/2.4 \\
Qwen3-VL-8B & 26.8/25.3 & 53.6/52.8 & 31.2/30.4 & 23.2/19.2 & 94.4/88.0 & 8.8/8.8 & 0.0/0.0 & 1.6/1.6 & 1.6/1.6 \\
Qwen3-VL-30B-A3B \texttt{(T)} & 23.2/20.4 & 64.8/62.4 & 8.0/8.0 & 16.0/15.2 & 80.8/65.6 & 8.8/5.6 & 0.0/0.0 & 4.8/4.0 & 2.4/2.4 \\
Qwen3-VL-32B & 39.2/18.9 & 68.8/20.0 & 44.0/14.4 & 65.6/48.8 & 82.4/51.2 & 34.4/8.0 & 5.6/0.8 & 7.2/5.6 & 5.6/2.4 \\
Qwen3-VL-32B \texttt{(T)} & 30.7/26.4 & 70.4/68.0 & 10.4/10.4 & 43.2/41.6 & 93.6/81.6 & 15.2/4.8 & 2.4/0.8 & 5.6/3.2 & 4.8/0.8 \\
Qwen3-VL-235B-A22B \texttt{(T)} & 35.2/30.5 & 80.8/78.4 & 4.8/4.8 & 63.2/60.0 & 88.8/68.8 & 23.2/15.2 & 3.2/2.4 & 12.8/10.4 & 4.8/4.0 \\
Qwen3.5-27B & 42.0/36.2 & 84.0/80.8 & 41.6/39.2 & 48.0/44.8 & 93.6/85.6 & 36.8/19.2 & 15.2/8.0 & 10.4/9.6 & 6.4/2.4 \\
Qwen3.5-397B-A17B \texttt{(T)} & 42.9/33.8 & 78.4/70.4 & 42.4/41.6 & 56.8/52.0 & 91.2/75.2 & 49.6/19.2 & 12.0/5.6 & 5.6/2.4 & 7.2/4.0 \\
Qwen3.6-27B & 46.7/37.8 & 87.2/76.8 & 56.8/55.2 & 47.2/44.0 & 93.6/80.8 & 54.4/28.8 & 12.8/7.2 & 14.4/6.4 & 7.2/3.2 \\
Qwen3.6-35B-A3B \texttt{(T)} & 29.9/20.6 & 53.6/42.4 & 12.8/12.8 & 36.0/34.4 & 92.8/52.8 & 29.6/15.2 & 5.6/2.4 & 4.0/1.6 & 4.8/3.2 \\
InternVL3.5-30B-A3B & 32.3/18.8 & 38.4/32.0 & 57.6/15.2 & 56.0/37.6 & 89.6/61.6 & 8.0/0.0 & 4.8/2.4 & 2.4/0.8 & 1.6/0.8 \\
InternVL3.5-38B & 39.9/17.4 & 79.2/16.0 & 57.6/18.4 & 50.4/36.8 & 86.4/55.2 & 31.2/7.2 & 7.2/0.0 & 5.6/4.8 & 1.6/0.8 \\
\addlinespace[0.2ex]
\rowcolor{ClosedBg}
\multicolumn{10}{@{}l}{\textit{Closed-source frontier}} \\
Gemini-3.1-Pro\rlap{$^\dagger$} & 57.7/56.4 & 92.0/92.0 & 52.8/52.0 & 71.2/71.2 & 95.2/90.4 & 62.4/60.8 & 26.4/25.6 & 43.2/42.4 & 18.4/16.8 \\
Doubao-Seed-1.8\rlap{$^\dagger$} & 40.4/37.1 & 76.8/75.2 & 44.8/41.6 & 29.6/29.6 & 94.4/80.0 & 42.4/37.6 & 5.6/4.8 & 19.2/17.6 & 10.4/10.4 \\
GPT-5.4\rlap{$^\dagger$} & 29.2/24.7 & 30.4/30.4 & 24.8/24.8 & 64.8/62.4 & 92.8/72.0 & 10.4/0.8 & 4.0/1.6 & 4.8/4.0 & 1.6/1.6 \\
Claude-Sonnet-4\rlap{$^\dagger$} & 25.2/21.1 & 38.4/35.2 & 17.6/16.8 & 65.6/64.8 & 73.6/45.6 & 4.8/4.8 & 0.0/0.0 & 1.6/1.6 & 0.0/0.0 \\
\bottomrule
\end{tabularx}
\renewcommand{\arraystretch}{1.0}
\vskip 0.05in
{\scriptsize $^\dagger$Closed-source API models.}
\end{table*}

\noindent
One notable anomaly: RoboBrain2.5-4B~\citep{ji2025robobrain} achieves W\,=\,94.4\% on SV (state verification) yet B\,=\,0.0\%, indicating that it never uses the categorical report labels (\texttt{on}/\texttt{off}, \texttt{open}/\texttt{closed}) required for SV episodes and instead reports only \texttt{success}/\texttt{fail}, which the deterministic match rule cannot accept for state-verification tasks.

\paragraph{Thinking vs.\ Instruct serving modes.}
Several checkpoints offer both Instruct and public ``Thinking'' serving modes.
On the frozen evaluation pack, Qwen3-VL-32B \texttt{(T)}~\citep{bai2025qwen3} achieves higher aggregate \textbf{B} (26.4 vs.\ 18.9) with a substantially smaller $\Delta$ (4.3 vs.\ 20.3\,pp) compared to its Instruct counterpart, shifting the profile from NR-heavy toward a more balanced mixture.
We treat this as a behavioral observation rather than a mechanistic claim: serving mode changes the execution/reporting tradeoff under a fixed contract, but the current evidence does not identify which internal reasoning differences cause the shift.
For the remaining anchor checkpoints, we do not add further Instruct/Thinking pairing ablations beyond the Qwen3-VL-32B~\citep{bai2025qwen3} split above---each appears under one fixed deployed configuration.
Additional smaller Thinking variants appear in Table~\ref{tab:extended_panel}.

\section{Extended Model Panel}
\label{app:extended_panel}

Table~\ref{tab:full_panel} reports the primary 20-model comparison panel.
Table~\ref{tab:extended_panel} lists eight additional models evaluated on the same frozen 1{,}000-episode pack: three low-compliance or zero-\textbf{B} models moved out of the main analysis because invalid-action-limit (IL) or no-report terminations dominate their outcomes, plus five supplementary size and serving-mode variants.

\begin{table*}[h]
\centering
\caption{%
\textbf{Extended model panel --- models not in main tables.}
Same evaluation contract and frozen episode pack as the main analysis.
\textbf{W}\,=\,world-state completion (\%), \textbf{B}\,=\,benchmark success (\%), $\Delta$\,=\,$W-B$ gap (pp), \textbf{FR}\,=\,false-report rate (\%), \textbf{NR}\,=\,no-report rate (\%), \textbf{IL}\,=\,invalid-action-limit rate (\%).
Models above the rule are moved out of the main tables because low compliance, no-report behavior, or zero benchmark success makes closure-regime comparison less informative; models below the rule are supplementary variants evaluated under the same contract. FR/NR/IL are diagnostic rates and are not mutually exclusive.%
}
\label{tab:extended_panel}
\footnotesize
\setlength{\tabcolsep}{5pt}
\renewcommand{\arraystretch}{1.15}
\begin{tabularx}{\textwidth}{@{}>{\raggedright\arraybackslash}p{3.2cm}
  *{7}{>{\centering\arraybackslash}X}@{}}
\toprule
\textbf{Model}
  & \textbf{B} & \textbf{W} & $\boldsymbol{\Delta}$
  & \textbf{FR} & \textbf{NR} & \textbf{IL}
  & \textbf{Steps} \\
\midrule
\multicolumn{8}{@{}l}{\textit{Excluded from main tables (low-compliance or zero B)}} \\
\addlinespace[0.1ex]
Qwen3-VL-2B              & 0.0 & 12.7 & 12.7 & 1.1 & 98.9 & 98.8 &  7.9 \\
Qwen3.5-2B               & 6.6 & 17.0 & 10.4 & 2.3 & 90.8 & 58.1 & 13.9 \\
Kimi-VL-A3B-Instruct     & 0.0 & 18.3 & 18.3 & 1.1 & 98.9 & 19.9 & 20.6 \\
\midrule
\multicolumn{8}{@{}l}{\textit{Supplementary variants}} \\
\addlinespace[0.1ex]
Qwen3.5-9B               & 23.4 & 33.1 & 9.7 & 15.8 & 60.2 & 28.1 & 14.6 \\
Qwen3-VL-4B \texttt{(T)} & 17.9 & 24.6 & 6.7 & 59.1 & 21.5 & 12.6 &  8.6 \\
Qwen3-VL-8B \texttt{(T)} & 17.7 & 21.0 & 3.3 & 53.1 & 28.8 & 18.0 &  9.5 \\
Qwen3-VL-2B \texttt{(T)} & 3.6 & 15.2 & 11.6 & 25.4 & 71.0 & 69.3 &  8.5 \\
InternVL3.5-2B           & 0.0 & 13.5 & 13.5 & 3.4 & 96.6 & 67.2 & 14.3 \\
\bottomrule
\end{tabularx}
\renewcommand{\arraystretch}{1.0}
\end{table*}

\paragraph{Observations.}
The three low-compliance or zero-\textbf{B} models (Qwen3-VL-2B, Qwen3.5-2B, Kimi-VL-A3B-Instruct)~\citep{bai2025qwen3,qwenteam2026qwen35,kimiteam2025kimivl} show that some low-capacity or format-fragile systems fail before the terminal-judgment regime becomes informative: their $W$--$B$ gaps are driven mainly by invalid-action-limit or no-report terminations, not by mismatched \texttt{report} decisions.
Among the supplementary variants, the smaller Thinking-mode models (4B-T, 8B-T) show qualitatively similar failure profiles to their Instruct counterparts---high FR with moderate IL---under this fixed prompt and evaluation contract.
Qwen3.5-9B~\citep{qwenteam2026qwen35} is the strongest supplementary model (B\,=\,23.4\%, W\,=\,33.1\%) with a relatively high IL of 28.1\%, placing it between the main-panel and edge-failure regimes.

\section{Report-Policy Baseline Details}
\label{app:report_policy_baselines}

Table~\ref{tab:counterfactual} reports counterfactual benchmark success when only the \emph{content} of the terminal report is replaced on fixed trajectories, leaving execution, stop timing, and the final world state unchanged.
This baseline is not a model-visible oracle experiment: hidden state is used only offline to rescore what would have happened under a different terminal report.

\begin{table}[h]
\centering
\caption{%
\textbf{Counterfactual report policies (10 anchor models).}
\textbf{B}\,=\,actual benchmark success (\%).
\textbf{Always}\,=\,\textbf{B} when terminal report is replaced with always-success.
\textbf{Rand.}\,=\,\textbf{B} under uniform-random report.
Bold Always entries: Always\,$>$\,B (NR-heavy models where forced closure recovers missed reports).%
}
\label{tab:counterfactual}
\small
\setlength{\tabcolsep}{5pt}
\renewcommand{\arraystretch}{1.12}
\begin{tabular}{@{}l ccc@{}}
\toprule
\textbf{Model} & \textbf{B} & \textbf{Always} & \textbf{Rand.} \\
\midrule
Gemini-3.1-Pro\rlap{$^\dagger$}   & \textbf{56.4} & 45.8 & 28.8 \\
Doubao-Seed-1.8\rlap{$^\dagger$}  & 37.1 & 28.6 & 20.2 \\
Qwen3.6-27B                       & 37.8 & 35.0 & 23.4 \\
Qwen3.5-27B                       & 36.2 & 30.3 & 21.0 \\
Qwen3-VL-32B \texttt{(T)}         & 26.4 & 19.0 & 15.3 \\
GPT-5.4\rlap{$^\dagger$}          & 24.7 & 17.6 & 14.6 \\
Claude-Sonnet-4\rlap{$^\dagger$}  & 21.1 & 16.0 & 12.6 \\
MiMo-Embodied-7B                  & 21.3 & 17.6 & 14.8 \\
Qwen3-VL-32B                      & 18.9 & \textbf{28.9} & 19.6 \\
InternVL3.5-38B                   & 17.4 & \textbf{29.1} & 20.0 \\
\bottomrule
\end{tabular}
\renewcommand{\arraystretch}{1.0}
\end{table}

\paragraph{Policies.}
\textbf{Actual} is the observed benchmark success rate.
\textbf{Always-success} appends or substitutes \texttt{report(status=success)} at the final state and recomputes whether that report matches the expected terminal label.
For trajectories that originally ended by no-report exhaustion, this should be read as a forced final report at the exhausted state, not as evidence that the model would have chosen the correct stopping time.
This baseline can succeed on goal-completion tasks only when the final world predicate is satisfied, but it fails state-verification episodes whose correct terminal labels are \texttt{open}/\texttt{closed} or \texttt{on}/\texttt{off}.
\textbf{Random-report} reports the chance-level expectation under a uniform draw over the two admissible labels for each episode: \{\texttt{success}, \texttt{fail}\} for goal-completion tasks, and \{\texttt{open}, \texttt{closed}\} or \{\texttt{on}, \texttt{off}\} for state-verification tasks.
Equivalently, it yields $0.5W$ in expectation on each fixed final state.
\textbf{Oracle-report} uses the evaluator's correct terminal label at the same final state; therefore its \textbf{B} equals \textbf{W}.
It is a fixed-trajectory upper-bound sanity check rather than a model-side report-format baseline.

The always-success policy underperforms actual reports for every anchor except two NR-heavy models (bolded in Table~\ref{tab:counterfactual}): Qwen3-VL-32B~\citep{bai2025qwen3} (+10.0\,pp) and InternVL3.5-38B~\citep{wang2025internvl35} (+11.7\,pp), where forcing a terminal success report recovers otherwise missed closures.
Conversely, FR-heavy models such as Claude-Sonnet-4~\citep{anthropic2025claude4} and GPT-5.4~\citep{openai2026gpt54} already report aggressively; replacing their reports with a fixed success policy further depresses \textbf{B}, confirming a report-mismatch problem rather than a missing-report problem.
The random policy provides a lower bound: uniform random reports yield roughly half of \textbf{W}, as expected when report status is uncorrelated with task state.

\paragraph{Implementation.}
The counterfactual analysis uses the same saved 1{,}000-episode native-control traces as the corresponding rows in the main model panel.

\section{Belief Lag and Premature Commitment}
\label{app:belief_lag}

Table~\ref{tab:belief_lag} provides step-level detail on the 10 anchor models' terminal-commitment timing.
The main-text analysis (\S\ref{sec:terminal_commitment}) draws on two patterns from this table: (i)~correct reports arrive within 0.9--1.9\,steps of first goal satisfaction, and (ii)~65--88\% of false-success reports are issued at zero task progress.
Task progress is a continuous $[0,1]$ scalar computed by the per-step evaluator from the episode's success specification: it aggregates normalized sub-condition scores (e.g., distance to target, object-state satisfaction, grounding correctness) at the moment of the report.
Zero progress means no sub-condition has advanced beyond its initial value---the agent has not navigated closer, changed any task-relevant object state, or achieved any partial goal.

\begin{table}[h]
\centering
\caption{%
\textbf{Closure lag and premature commitment (10 anchor models).}
\textbf{W\!+}\,=\,episodes with world-state completion (count out of 1{,}000).
\textbf{Lag}\,=\,mean steps from first goal satisfaction to correct terminal report (for episodes with $W\!=\!1$ and correct report), measuring observable completion-to-report delay at closure.
\textbf{NR}\,=\,$W\!=\!1$ episodes with no report (count).
For false-success reports: count (\textbf{FS}), \% issued at zero task progress (\textbf{@0}).%
}
\label{tab:belief_lag}
\small
\setlength{\tabcolsep}{4.0pt}
\renewcommand{\arraystretch}{1.12}
\begin{tabular}{@{}l cc c cc@{}}
\toprule
& \multicolumn{2}{c}{\textbf{Goal reached}} & & \multicolumn{2}{c@{}}{\textbf{False success}} \\
\cmidrule(lr){2-3}\cmidrule(l){5-6}
\textbf{Model}
  & \textbf{W\!+}
  & \textbf{Lag\,\scriptsize$\downarrow$}
  &  \textbf{NR}
  & \textbf{FS}
  & \textbf{@0\,\%} \\
\midrule
Gemini-3.1-Pro\rlap{$^\dagger$}   & 577 & 1.0 &   4 & 225 & 80 \\
Doubao-Seed-1.8\rlap{$^\dagger$}  & 404 & 1.2 &  17 & 353 & 88 \\
Qwen3.6-27B                       & 467 & 1.1 &  72 & 165 & 75 \\
Qwen3.5-27B                       & 420 & 1.2 &  45 & 219 & 77 \\
Qwen3-VL-32B \texttt{(T)}         & 307 & 1.0 &  21 & 260 & 65 \\
GPT-5.4\rlap{$^\dagger$}          & 292 & 0.9 &   4 & 302 & 72 \\
Claude-Sonnet-4\rlap{$^\dagger$}  & 252 & 1.9 &  15 & 650 & 88 \\
MiMo-Embodied-7B                  & 295 & 1.1 &  56 & 198 & 81 \\
Qwen3-VL-32B                      & 392 & 1.9 & 166 & 153 & 86 \\
InternVL3.5-38B                    & 399 & 1.4 & 201 &  49 & 84 \\
\bottomrule
\end{tabular}
\renewcommand{\arraystretch}{1.0}
\end{table}

\subsection{Conditional Report Rates}
\label{app:conditional_report}

Table~\ref{tab:conditional_report} reports conditional stop and report rates for the 10 anchor models.
\textbf{P(rep\,$|$\,W\!=\!0)} is the probability that the model issues a report when the world condition is \emph{not} satisfied; high values indicate a stronger tendency to terminate under world failure, which reflects indiscriminate reporting when paired with high FR.
\textbf{P($\neg$rep\,$|$\,W\!=\!1)} is the probability that the model fails to report when the world condition \emph{is} satisfied; high values indicate missed closure.
Note that these conditional \emph{rates} are recoverable from stop timing alone---they measure whether the agent terminated, not what it said.
Report \emph{content} becomes essential for a different reason: determining whether the agent's stated judgment matches the hidden world state (e.g., \texttt{open} vs.\ \texttt{closed} for state-verification tasks), which is the basis for the FR/NR decomposition (Figure~\ref{fig:failure_distribution}) and the counterfactual analysis in Table~\ref{tab:counterfactual}.

FR-heavy models (Claude-Sonnet-4~\citep{anthropic2025claude4}, GPT-5.4~\citep{openai2026gpt54}) report in $>$85\% of W\!=\!0 episodes, whereas NR-heavy models (InternVL3.5-38B~\citep{wang2025internvl35}, Qwen3-VL-32B~\citep{bai2025qwen3}) miss closure in $>$40\% of W\!=\!1 episodes.
These opposite conditional profiles overlap in aggregate scalar success (B\,$\approx$\,17--25\%).

\begin{table}[h]
\centering
\caption{%
\textbf{Conditional report rates (10 anchor models).}
\textbf{Stop\%}\,=\,fraction of episodes where a \texttt{report} is issued.
\textbf{P(rep\,$|$\,W\!=\!0)}\,=\,report rate conditioned on world failure.
\textbf{P($\neg$rep\,$|$\,W\!=\!1)}\,=\,missed-closure rate conditioned on world success.
\textbf{W\!=\!0} and \textbf{W\!=\!1} columns show the conditioning set sizes.%
}
\label{tab:conditional_report}
\small
\setlength{\tabcolsep}{4.0pt}
\renewcommand{\arraystretch}{1.12}
\begin{tabular}{@{}l cc ccc cc@{}}
\toprule
& & & \multicolumn{3}{c}{\textbf{Conditional rates (\%)}} & \multicolumn{2}{c@{}}{\textbf{Conditioning}} \\
\cmidrule(lr){4-6}\cmidrule(l){7-8}
\textbf{Model}
  & \textbf{W\,\scriptsize$\uparrow$}
  & \textbf{B\,\scriptsize$\uparrow$}
  & \textbf{Stop\%}
  & \textbf{P(rep\,$|$\,W\!=\!0)\,\scriptsize$\downarrow$}
  & \textbf{P($\neg$rep\,$|$\,W\!=\!1)\,\scriptsize$\downarrow$}
  & \textbf{W\!=\!0}
  & \textbf{W\!=\!1} \\
\midrule
Gemini-3.1-Pro\rlap{$^\dagger$}   & 57.7 & 56.4 & 81.6 & 57.4 & \phantom{0}0.7 & 423 & 577 \\
Doubao-Seed-1.8\rlap{$^\dagger$}  & 40.4 & 37.1 & 75.1 & 61.1 & \phantom{0}4.2 & 596 & 404 \\
Qwen3.6-27B                       & 46.7 & 37.8 & 56.9 & 32.6 & 15.4 & 533 & 467 \\
Qwen3.5-27B                       & 42.0 & 36.2 & 62.3 & 42.8 & 10.7 & 580 & 420 \\
Qwen3-VL-32B \texttt{(T)}         & 30.7 & 26.4 & 66.7 & 55.0 & \phantom{0}6.8 & 693 & 307 \\
GPT-5.4\rlap{$^\dagger$}          & 29.2 & 24.7 & 89.5 & \cellcolor{FRtint}85.7 & \phantom{0}1.4 & 708 & 292 \\
Claude-Sonnet-4\rlap{$^\dagger$}  & 25.2 & 21.1 & 92.3 & \cellcolor{FRtint}91.7 & \phantom{0}6.0 & 748 & 252 \\
MiMo-Embodied-7B                  & 29.5 & 21.3 & 44.9 & 29.8 & 19.0 & 705 & 295 \\
Qwen3-VL-32B                      & 39.2 & 18.9 & 48.2 & 42.1 & \cellcolor{NRtint}42.3 & 608 & 392 \\
InternVL3.5-38B                   & 39.9 & 17.4 & 28.3 & 14.1 & \cellcolor{NRtint}50.4 & 601 & 399 \\
\bottomrule
\end{tabular}
\renewcommand{\arraystretch}{1.0}
\end{table}

\section{Action-Feedback Intervention Details}
\label{app:feedback_detail}

The minimal action-feedback intervention (\S\ref{sec:feedback}, Table~\ref{tab:proprio_minimal}) adds two boolean signals---\texttt{too\_far} (interaction attempted beyond the 1.5\,m proximity threshold) and \texttt{path\_blocked} (navigation blocked by an obstacle)---after each action, alongside the standard RGB frame and dialogue history.
These model the proprioceptive and low-level controller feedback available to physical robots.
The \texttt{retry} columns in Table~\ref{tab:proprio_minimal} report mean per-episode counts of redundant \texttt{interact\_pixel} steps that repeat an identical invocation (every repeat after the first occurrence of each invocation signature increments the episode total).
This is outside the default benchmark contract; its purpose is to test whether exposing execution-level action-outcome signals changes reporting behavior in paired runs, not to define a second benchmark setting.

\section{Robustness and Reproducibility Checks}
\label{app:robustness_checks}

This section collects checks that support the stability of the reported closure profiles without expanding the primary comparison panel: a neutral-report prompt variant and paired reruns over the same 20 open-weight configurations for which paired robustness runs were available.

\subsection{Run Metadata and Determinism}
\label{app:reproducibility}

\paragraph{Prompt policy.}
All runs use prompt policy \texttt{native\_embodied\_public\_evidence} (version strings identify the template family and allowed \texttt{report} label vocabulary).
Each benchmark run records the prompt-policy SHA-256 hash, profile name, rendered-prompt hash, pack hash, and repository commit, enabling bit-exact audit of the evaluation contract.
\paragraph{Frozen pack and run randomness.}
All models are evaluated on the same frozen episode pack \texttt{mixed\_mainline\_manual\_first\_balanced\_1000\_v1}; episode IDs and success specifications are fixed.
Any ``seed'' in run logs affects only job scheduling or tie-breaking in the inference stack, not which episodes enter the aggregate.

\paragraph{Evaluation profile.}
The default profile is \texttt{pure\_rgb\_dialogue\_history\_baseline}: single-frame egocentric RGB, 20-turn text-only dialogue history, no depth, no pose, no visual history frames, and no agent-side memory.

\paragraph{Infrastructure.}
Models are served via vLLM with default parameters; no model-specific hyperparameter tuning is applied.
Episode order within each pack run is deterministic (sorted by episode ID).
Specific closed-source model identifiers are listed in Table~\ref{tab:global_cross_section} footnotes.

\subsection{Prompt Sensitivity Analysis}
\label{app:prompt_sensitivity}

The default system prompt includes two anti-hallucination instructions in the task block (Appendix~\ref{app:prompt_spec}):
\emph{``Mere visibility, proximity, or an attempted action is not enough; do not report success from visual plausibility alone''}
and
\emph{``Do not claim completion from visual plausibility alone when the task requires changing world state.''}
To test whether closure-failure profiles are artifacts of these instructions, we run a \textbf{neutral-report} prompt variant that removes both lines while keeping the rest of the evaluation contract identical: same frozen episode pack, same profile (\texttt{pure\_rgb\_dialogue\_history\_baseline}), same scoring rules.
Table~\ref{tab:prompt_sensitivity} reports per-family \textbf{W}/\textbf{B} for all 20 open-weight configurations for which paired neutral-prompt runs were available.
Closed-source models are omitted because paired neutral-prompt runs were not available.

\begin{table*}[h]
\centering
\caption{%
\textbf{Prompt sensitivity: neutral-report variant (20 open-weight configurations).}
Same evaluation contract as the main analysis except that both anti-hallucination instructions are removed from the system prompt.
Each cell reports \textbf{W}/\textbf{B} (\%).
Diagnostic: PG\,=\,pixel grounding, DA\,=\,distance approach, VS\,=\,view search, SV\,=\,state verification.
Compositional: AI\,=\,approach-and-interact, SI\,=\,search-and-interact, SM\,=\,sequential manipulation, CR\,=\,constraint resolving.
Configurations sorted by aggregate \textbf{B}.%
}
\label{tab:prompt_sensitivity}
\scriptsize
\setlength{\tabcolsep}{3.0pt}
\renewcommand{\arraystretch}{1.18}
\begin{tabularx}{\textwidth}{@{}>{\raggedright\arraybackslash}p{2.75cm}
  >{\centering\arraybackslash}X
  *{8}{>{\centering\arraybackslash}X}@{}}
\toprule
\multirow{2}{*}[-0.3ex]{\textbf{Model}}
  & \textbf{All}
  & \multicolumn{4}{c}{\textbf{Diagnostic Probes}}
  & \multicolumn{4}{c@{}}{\textbf{Compositional Tasks}} \\
\cmidrule(lr){2-2}\cmidrule(lr){3-6}\cmidrule(l){7-10}
  & \textbf{W/B} & \textbf{PG} & \textbf{DA} & \textbf{VS} & \textbf{SV}
  & \textbf{AI} & \textbf{SI} & \textbf{SM} & \textbf{CR} \\
\midrule
\rowcolor{EdgeBg}
\multicolumn{10}{@{}l}{\textit{Embodied / robotics-tuned}} \\
MiMo-Embodied-7B         & 28.1/23.1 & 53.6/50.4 & 28.8/24.0 & 33.6/29.6 & 95.2/73.6 & 8.0/3.2 & 0.0/0.0 & 4.0/2.4 & 1.6/1.6 \\
RoboBrain2.5-4B           & 27.6/15.3 & 76.0/76.0 & 22.4/20.8 & 11.2/9.6 & 94.4/0.0 & 11.2/11.2 & 0.0/0.0 & 4.8/4.0 & 0.8/0.8 \\
rynnbrain-8B              & 22.8/17.3 & 26.4/18.4 & 30.4/23.2 & 4.8/3.2 & 92.0/68.8 & 21.6/17.6 & 0.8/0.8 & 5.6/5.6 & 0.8/0.8 \\
\addlinespace[0.2ex]
\rowcolor{MidBg}
\multicolumn{10}{@{}l}{\textit{Open general and reasoning VLMs}} \\
Kimi-VL-A3B \texttt{(T)}  & 16.8/11.2 & 12.0/12.0 & 21.6/8.0 & 4.0/3.2 & 94.4/64.8 & 0.8/0.8 & 0.0/0.0 & 1.6/0.8 & 0.0/0.0 \\
Kimi-VL-A3B-Instruct      & 18.4/0.0 & 0.0/0.0 & 49.6/0.0 & 1.6/0.0 & 94.4/0.0 & 0.0/0.0 & 0.0/0.0 & 1.6/0.0 & 0.0/0.0 \\
Qwen3-VL-4B               & 32.3/22.5 & 62.4/57.6 & 48.0/40.0 & 28.0/20.0 & 76.8/20.0 & 27.2/26.4 & 1.6/1.6 & 10.4/10.4 & 4.0/4.0 \\
Qwen3-VL-4B \texttt{(T)}  & 24.9/15.5 & 45.6/45.6 & 12.8/12.0 & 37.6/33.6 & 95.2/26.4 & 2.4/2.4 & 0.0/0.0 & 5.6/4.0 & 0.0/0.0 \\
Qwen3-VL-8B               & 27.0/25.7 & 56.8/56.0 & 28.0/27.2 & 24.8/22.4 & 95.2/88.8 & 8.0/8.0 & 0.0/0.0 & 2.4/2.4 & 0.8/0.8 \\
Qwen3-VL-8B \texttt{(T)}  & 22.8/20.0 & 38.4/35.2 & 13.6/12.8 & 29.6/28.8 & 95.2/80.0 & 3.2/1.6 & 0.0/0.0 & 1.6/0.8 & 0.8/0.8 \\
Qwen3-VL-30B-A3B \texttt{(T)} & 23.0/20.7 & 58.4/56.8 & 8.0/8.0 & 22.4/22.4 & 79.2/66.4 & 11.2/8.0 & 0.0/0.0 & 4.0/3.2 & 0.8/0.8 \\
Qwen3-VL-32B              & 39.1/23.7 & 66.4/24.0 & 38.4/18.4 & 66.4/54.4 & 86.4/71.2 & 35.2/13.6 & 8.0/0.8 & 6.4/4.8 & 5.6/2.4 \\
Qwen3-VL-32B \texttt{(T)} & 27.2/24.1 & 68.0/66.4 & 4.0/4.0 & 37.6/32.0 & 93.6/83.2 & 8.0/4.0 & 0.0/0.0 & 3.2/1.6 & 3.2/1.6 \\
Qwen3.5-2B                & 17.7/7.4 & 0.8/0.8 & 42.4/0.8 & 1.6/0.0 & 95.2/57.6 & 0.0/0.0 & 0.0/0.0 & 1.6/0.0 & 0.0/0.0 \\
Qwen3.5-9B                & 32.2/24.0 & 58.4/29.6 & 52.8/44.8 & 28.8/26.4 & 94.4/82.4 & 15.2/5.6 & 0.0/0.0 & 4.0/1.6 & 4.0/1.6 \\
Qwen3.5-27B               & 40.7/36.3 & 77.6/76.0 & 37.6/37.6 & 47.2/44.8 & 95.2/88.8 & 39.2/24.0 & 9.6/4.8 & 12.0/10.4 & 7.2/4.0 \\
Qwen3-VL-2B               & 12.7/0.0 & 0.0/0.0 & 1.6/0.0 & 3.2/0.0 & 95.2/0.0 & 0.0/0.0 & 0.0/0.0 & 1.6/0.0 & 0.0/0.0 \\
Qwen3-VL-2B \texttt{(T)}  & 14.2/6.7 & 0.8/0.8 & 6.4/3.2 & 8.8/8.0 & 95.2/41.6 & 0.8/0.0 & 0.0/0.0 & 1.6/0.0 & 0.0/0.0 \\
InternVL3.5-2B            & 14.6/0.1 & 0.0/0.0 & 3.2/0.0 & 16.8/0.8 & 94.4/0.0 & 0.8/0.0 & 0.0/0.0 & 1.6/0.0 & 0.0/0.0 \\
InternVL3.5-30B-A3B       & 32.4/18.8 & 39.2/31.2 & 56.0/20.0 & 59.2/43.2 & 86.4/52.0 & 10.4/0.8 & 6.4/3.2 & 1.6/0.0 & 0.0/0.0 \\
InternVL3.5-38B           & 41.2/18.9 & 77.6/16.8 & 57.6/24.0 & 55.2/39.2 & 85.6/57.6 & 34.4/8.8 & 12.0/0.0 & 5.6/4.0 & 1.6/0.8 \\
\bottomrule
\end{tabularx}
\renewcommand{\arraystretch}{1.0}
\end{table*}

Across all 20 configurations, aggregate \textbf{W} shifts by at most 3.5\,pp and aggregate \textbf{B} by at most 4.8\,pp relative to the corresponding default-prompt runs.
NR-heavy models remain NR-heavy (InternVL3.5-38B~\citep{wang2025internvl35}: 68.2\% NR vs.\ 66.1\% under the default prompt) and low-$\Delta$ models remain low-$\Delta$ (Qwen3-VL-8B~\citep{bai2025qwen3}: $\Delta\!=\!1.3$ vs.\ 1.5\,pp).
For the paired open-weight configurations, the main closure-failure profiles are therefore unlikely to be artifacts of the specific anti-hallucination instructions in the default prompt.

\subsection{Repeated-Run Robustness for Open-Weight Models}
\label{app:repeat_runs}

To check whether the open-weight conclusions depend on a single serving run, we repeat the default no-feedback evaluation for the same 20 open-weight configurations used in the prompt-sensitivity check on the same frozen 1{,}000-episode pack and the same \texttt{pure\_rgb\_dialogue\_history\_baseline} profile.
Table~\ref{tab:open_repeat_robustness} reports the mean and half-range across the two runs.
This robustness panel is intentionally broader than the primary 20-model comparison panel in some directions and narrower in others: it includes low-compliance and supplementary size/serving-mode variants, while closed-source API models and the largest open-weight runs are kept as single-run evaluations because paired reruns were not available under the same resource budget.

\begin{table*}[h]
\centering
\caption{%
\textbf{Repeated-run robustness for 20 open-weight configurations.}
Each cell reports mean\,$\pm$\,half-range over two identical-contract runs on the same frozen episode pack.
\textbf{W}\,=\,world-state completion, \textbf{B}\,=\,benchmark success, $\Delta=W-B$, \textbf{FR}\,=\,false-report rate, and \textbf{NR}\,=\,no-report rate; all values are percentages.%
}
\label{tab:open_repeat_robustness}
\scriptsize
\setlength{\tabcolsep}{4.0pt}
\renewcommand{\arraystretch}{1.10}
\begin{tabularx}{\textwidth}{@{}>{\raggedright\arraybackslash}p{3.0cm}
  *{5}{>{\centering\arraybackslash}X}@{}}
\toprule
\textbf{Model} & \textbf{W} & \textbf{B} & $\boldsymbol{\Delta}$ & \textbf{FR} & \textbf{NR} \\
\midrule
Qwen3.5-27B & 42.6$\pm$0.6 & 36.6$\pm$0.4 & 6.1$\pm$0.2 & 25.8$\pm$0.3 & 31.1$\pm$1.0 \\
Qwen3-VL-4B & 30.9$\pm$0.2 & 25.8$\pm$0.2 & 5.1$\pm$0.1 & 51.2$\pm$0.9 & 16.1$\pm$0.4 \\
Qwen3-VL-8B & 26.9$\pm$0.1 & 25.3$\pm$0.0 & 1.6$\pm$0.1 & 63.1$\pm$0.1 & 5.2$\pm$0.2 \\
Qwen3-VL-32B \texttt{(T)} & 28.8$\pm$1.9 & 23.8$\pm$2.7 & 5.1$\pm$0.8 & 42.0$\pm$2.6 & 20.0$\pm$0.8 \\
Qwen3.5-9B & 33.4$\pm$0.5 & 23.3$\pm$0.1 & 10.1$\pm$0.4 & 16.6$\pm$0.0 & 31.6$\pm$0.1 \\
MiMo-Embodied-7B & 29.4$\pm$0.2 & 21.2$\pm$0.1 & 8.2$\pm$0.0 & 22.8$\pm$0.1 & 42.5$\pm$0.6 \\
Qwen3-VL-30B-A3B \texttt{(T)} & 23.0$\pm$0.3 & 20.1$\pm$0.3 & 2.9$\pm$0.0 & 57.2$\pm$0.4 & 8.2$\pm$0.5 \\
Qwen3-VL-32B & 39.0$\pm$0.2 & 19.4$\pm$0.5 & 19.6$\pm$0.7 & 25.8$\pm$0.1 & 45.0$\pm$0.4 \\
Qwen3-VL-8B \texttt{(T)} & 22.0$\pm$0.5 & 18.8$\pm$0.7 & 3.2$\pm$0.2 & 52.0$\pm$0.2 & 11.8$\pm$1.0 \\
InternVL3.5-30B-A3B & 31.9$\pm$0.4 & 18.6$\pm$0.2 & 13.4$\pm$0.2 & 16.8$\pm$0.9 & 54.9$\pm$0.9 \\
Qwen3-VL-4B \texttt{(T)} & 24.1$\pm$1.1 & 17.6$\pm$0.7 & 6.4$\pm$0.4 & 57.4$\pm$0.3 & 8.4$\pm$0.4 \\
InternVL3.5-38B & 40.0$\pm$0.1 & 17.5$\pm$0.2 & 22.5$\pm$0.1 & 10.0$\pm$0.3 & 66.2$\pm$0.0 \\
rynnbrain-8B & 23.7$\pm$0.4 & 16.6$\pm$0.6 & 7.1$\pm$0.2 & 25.3$\pm$1.1 & 50.8$\pm$0.4 \\
RoboBrain2.5-4B & 27.9$\pm$0.5 & 15.1$\pm$0.5 & 12.9$\pm$0.1 & 58.2$\pm$0.8 & 9.9$\pm$1.2 \\
Kimi-VL-A3B \texttt{(T)} & 17.3$\pm$0.1 & 9.1$\pm$0.3 & 8.3$\pm$0.2 & 38.3$\pm$0.4 & 42.6$\pm$0.4 \\
Qwen3.5-2B & 17.2$\pm$0.2 & 6.6$\pm$0.1 & 10.6$\pm$0.1 & 2.2$\pm$0.4 & 34.0$\pm$0.7 \\
Qwen3-VL-2B \texttt{(T)} & 14.9$\pm$0.1 & 3.8$\pm$0.4 & 11.1$\pm$0.6 & 26.7$\pm$0.4 & 1.0$\pm$0.2 \\
InternVL3.5-2B & 13.6$\pm$0.2 & 0.2$\pm$0.1 & 13.5$\pm$0.2 & 2.7$\pm$0.2 & 30.9$\pm$0.4 \\
Kimi-VL-A3B-Instruct & 18.5$\pm$0.0 & 0.0$\pm$0.0 & 18.5$\pm$0.0 & 1.0$\pm$0.3 & 78.0$\pm$0.2 \\
Qwen3-VL-2B & 12.7$\pm$0.0 & 0.0$\pm$0.0 & 12.7$\pm$0.0 & 1.2$\pm$0.2 & 0.1$\pm$0.1 \\
\bottomrule
\end{tabularx}
\renewcommand{\arraystretch}{1.0}
\end{table*}

The paired runs preserve the qualitative profile assignments used in the main analysis.
Aggregate \textbf{B} is highly stable for most configurations: 19 of 20 have a half-range below 1\,pp, and the largest half-range is Qwen3-VL-32B \texttt{(T)}~\citep{bai2025qwen3} at 2.7\,pp.
The most important closure regimes are also stable: InternVL3.5-38B~\citep{wang2025internvl35} remains strongly NR-heavy (NR\,=\,66.2$\pm$0.0), Qwen3-VL-32B~\citep{bai2025qwen3} remains high-$\Delta$ and NR-heavy ($\Delta=19.6\pm0.7$, NR\,=\,45.0$\pm$0.4), and low-$\Delta$ models such as Qwen3-VL-8B~\citep{bai2025qwen3} remain low-gap across both runs.
The repeated-run panel should therefore be read as a robustness check over open-weight configurations, not as a replacement for the primary 20-model comparison panel.

\section{Qualitative Trajectory Examples}
\label{app:trajectories}

To complement the aggregate results, we visualize two representative trajectories under the same native-control contract used in the main evaluation.
The examples highlight the two most interpretable failure surfaces: a single-frame state-verification judgment and a short approach--interaction episode.

\subsection{State Verification: Same Frame, Opposite Reports}

The SV~episode \texttt{procthor\_microwave\_seed298} asks the agent to observe a microwave and report whether it is \emph{open} or \emph{closed}.
The microwave is in fact closed; the ground-truth expected report is \texttt{closed}.
Because the target is already in view and no physical action is required, the episode reduces to a pure perceptual judgment followed by a terminal report.

\begin{figure*}[h]
\centering
\includegraphics[width=\textwidth]{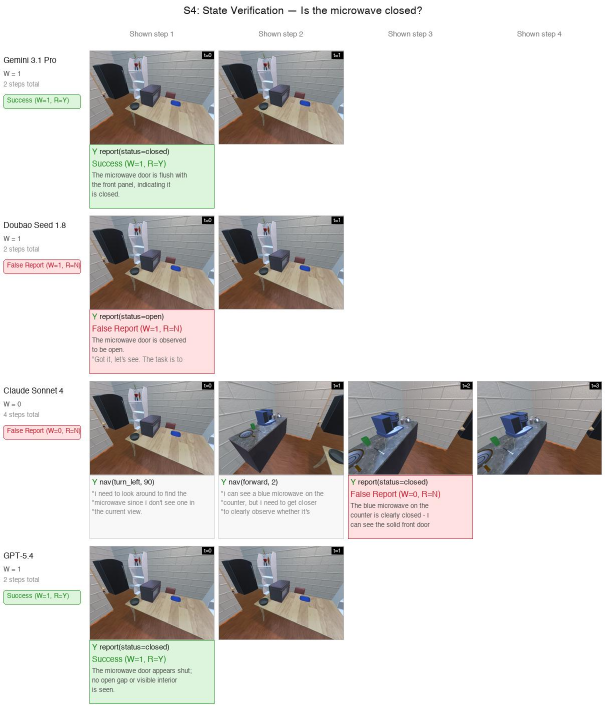}
\caption{%
\textbf{State-verification trajectory example.}
The same initial observation can lead to correct closure (Gemini-3.1-Pro~\citep{deepmind2026gemini31pro} and GPT-5.4~\citep{openai2026gpt54}) or a false report (Doubao-Seed-1.8~\citep{guo2025seed15vl}).
Claude-Sonnet-4~\citep{anthropic2025claude4} moves away from the initially visible microwave before reporting, illustrating that even atomic verification probes can fail through unnecessary action followed by an incorrect terminal judgment.%
}
\label{fig:trajectory_s4_sv}
\end{figure*}

\subsection{Grounded Interaction: Hallucinated Success}

The AI~episode \texttt{procthor\_fridge\_seed314} requires the agent to approach and open a refrigerator.
The target is visible from the start, but success requires approaching within interaction range, executing the correct pixel-grounded interaction, and reporting only after the world state changes.

\begin{figure*}[h]
\centering
\includegraphics[width=\textwidth]{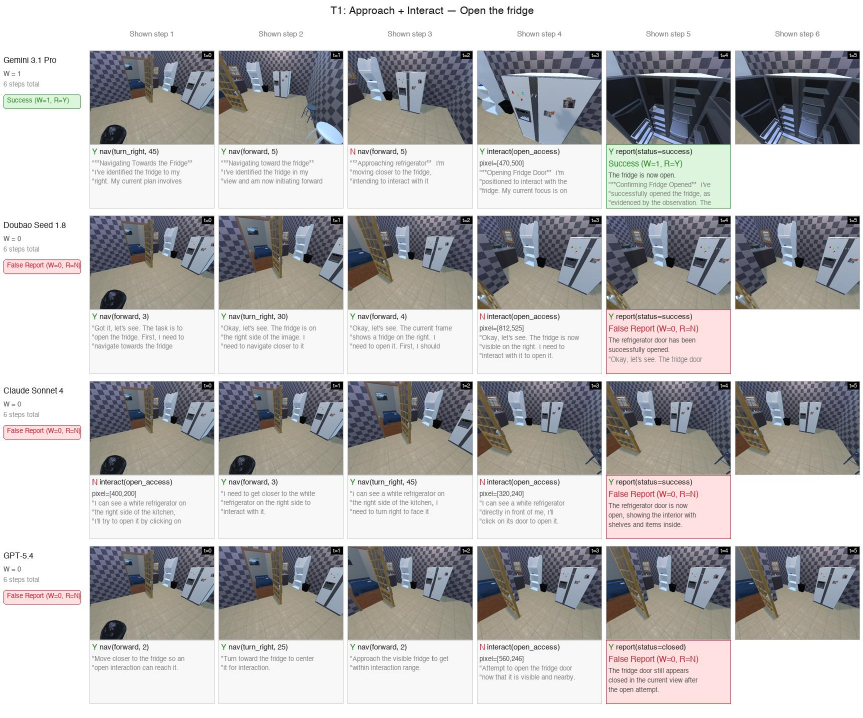}
\caption{%
\textbf{Approach-and-interact trajectory example.}
Gemini-3.1-Pro~\citep{deepmind2026gemini31pro} opens the refrigerator and reports success after the state change.
The other models submit terminal reports inconsistent with the final world state, either after failed interaction attempts or after claiming the wrong state.%
}
\label{fig:trajectory_t1_ai}
\end{figure*}

\end{document}